\documentclass{article}

\usepackage{microtype}
\usepackage{graphicx}
\usepackage{subcaption}
\usepackage{booktabs} %
\usepackage{amsmath}
\usepackage{amssymb}
\usepackage{mathtools}
\usepackage{amsthm}
\usepackage{multirow}
\usepackage{makecell}
\usepackage{overpic}
\usepackage{hyperref}
\usepackage[capitalize,noabbrev]{cleveref}

\theoremstyle{plain}

\theoremstyle{definition}

\theoremstyle{remark}

\usepackage[textsize=tiny]{todonotes}

\usepackage[accepted]{icml2026}

\definecolor{cvprblue}{rgb}{0.21,0.49,0.74}

\icmltitlerunning{FunPhase: A Periodic Functional Autoencoder for  Motion Generation via Phase Manifolds}

\begin{document}

\twocolumn[
  \icmltitle{FunPhase: A Periodic Functional Autoencoder for  Motion Generation \\ via Phase Manifolds}

  \icmlsetsymbol{equal}{*}
    \icmlsetsymbol{intern}{$\dagger$}

  \begin{icmlauthorlist}
    \icmlauthor{Marco Pegoraro}{intern,ISTA,autodesk}
    \icmlauthor{Evan Atherton}{autodesk}
    \icmlauthor{Bruno Roy}{autodesk}
    \icmlauthor{Aliasghar Khani}{autodesk}
    \icmlauthor{Arianna Rampini}{autodesk}
  \end{icmlauthorlist}

  \icmlaffiliation{autodesk}{Autodesk Research}
  \icmlaffiliation{ISTA}{Institute of Science and Technology Austria}

  \icmlcorrespondingauthor{Marco Pegoraro}{marco.pegoraro@ist.ac.at}
  \icmlcorrespondingauthor{Arianna Rampini}{arianna.rampini@autodesk.com}

  \icmlkeywords{Motion Generation, Phase Manifold}

\vskip 0.3in

\begingroup
\setkeys{Gin}{width=0.95\linewidth}
\begin{center}
  \includegraphics{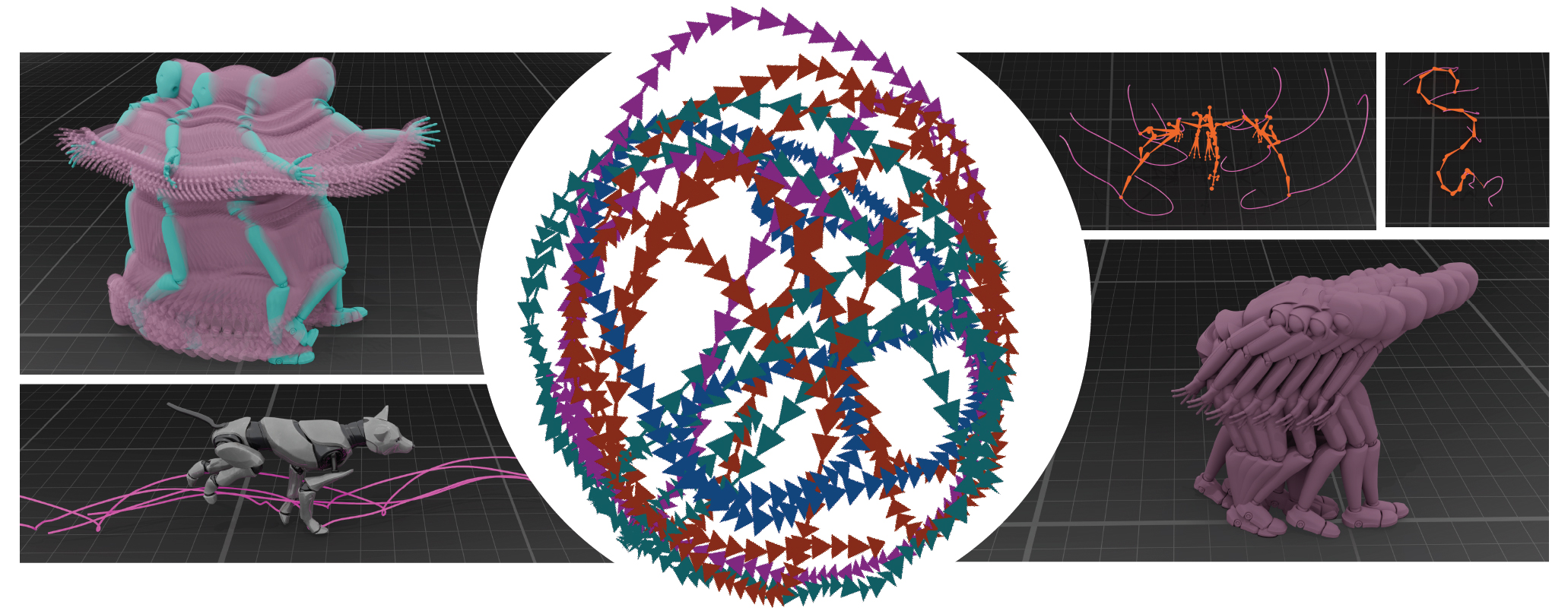}
\end{center}
\captionof{figure}{FunPhase is a functional periodic autoencoder that learns a phase-structured manifold for motion, enabling smooth, continuous spatio-temporal reconstruction, and skeleton-agnostic motion prediction and generation.}
\label{fig:teaser}
\endgroup

\vskip 0.2in

]

\printAffiliationsAndNotice{\textsuperscript{$\dagger$}Work done during an internship at Autodesk.\ }

\begin{abstract}
Learning natural body motion remains challenging due to the strong coupling between spatial geometry and temporal dynamics.
Embedding motion in phase manifolds, latent spaces that capture local periodicity, has proven effective for motion prediction; however, existing approaches are tied to fixed skeletons and narrow motion distributions, limiting their applicability across diverse settings.
We introduce FunPhase, a functional periodic autoencoder that learns a phase manifold for motion and replaces discrete temporal decoding with a function-space formulation, enabling smooth trajectories that can be sampled at arbitrary temporal resolutions.
FunPhase unifies motion prediction and generation within a single interpretable phase manifold, enabling motion generation via latent diffusion, generalizes across skeletons and datasets, and supports downstream tasks such as motion super-resolution and partial-body completion.
Our model achieves substantially lower reconstruction error than prior periodic autoencoder baselines, achieving uniform improvements of at least $45\%$ across all metrics, while enabling a broader range of applications and performing on par with state-of-the-art motion generation methods.
\end{abstract}
 
\section{Introduction}
\label{sec:intro}

Motion generation refers to the task of synthesizing realistic and coherent sequences of human or character movements, typically represented as 3D joint trajectories or poses, based on diverse inputs such as action labels, textual descriptions, or environmental cues. This capability is central to computer vision, computer graphics, robotics, and human-computer interaction, with applications spanning virtual avatar animation, video games, and embodied agents \cite{khani2025motion}. Effective motion generation automates traditionally manual and time-intensive animation processes, while also enabling behavioral modeling, simulation, and data augmentation.

Despite recent progress in generative modeling, learning motion remains a challenging problem due to the complex coupling between spatial geometry and temporal dynamics \cite{wang2019spatio}.  
Generic diffusion or autoregressive models that excel in visual and language domains often fail to produce physically plausible results.
The difficulty arises from the sparsity and highly nonlinear structure of the motion space, which can lead to artifacts such as root drift, inconsistent contacts, or temporal jitter \cite{zhu2023human}.

A promising direction to address these issues is the use of \emph{phase representations}, which encode the temporal progression of motion in a compact, interpretable form~\cite{starke2022deepphase}.
Recent work, such as \textit{DeepPhase} \cite{starke2022deepphase}, shows that phase-aware representations significantly improve motion alignment and prediction.
However, existing models remain restricted to fixed skeletons and narrow motion distributions, and are not easily extensible to probabilistic or generative frameworks.

Our idea is to overcome the limitations of frame-based motion representations by formulating motion as a continuous spatio-temporal function. 
We introduce \textbf{FunPhase}, a \emph{functional periodic autoencoder} that learns a phase manifold for motion while reconstructing motion as a continuous spatio-temporal function rather than a discrete frame sequence.
This functional formulation not only produces smooth trajectories that can be sampled at arbitrary temporal resolutions but also provides a natural interface for integrating physics priors \cite{wang2025fundiff}, which are of fundamental importance in the motion domain \cite{yuan2022physdiff}.
Moreover, it scales across heterogeneous skeletons and unifies motion prediction and generation within a single interpretable framework.
Our architecture builds upon the functional generative framework \cite{wang2025fundiff}, while preserving the interpretability of phase decomposition introduced in DeepPhase~\cite{starke2022deepphase}.

The main contributions of this work are summarized as follows:
\begin{enumerate}
   \item We propose \textbf{FunPhase}, a function-space autoencoder for motion that reconstructs movement as a continuous spatio-temporal function, enabling skeleton-agnostic encoding, arbitrary-resolution decoding, and a unified treatment of variable-rate temporal and partial spatial inputs. To our knowledge, this is the first motion model to operate in function space.
    \item As a complementary inductive bias, we integrate a periodic decomposition of the latent space and show that it acts as an effective regularizer, improving physical plausibility in reconstruction and denoising stability in diffusion-based generation.
    \item We demonstrate the framework's versatility across motion control, super-resolution, and partial-body completion, and use the function-space latents for class-conditional generation via latent diffusion, matching state-of-the-art performance.
\end{enumerate}

Overall, FunPhase bridges structured kinematic modeling with modern generative methods, providing a unified, skeleton-agnostic framework for learning, predicting, and generating motion functions.

\section{Related Work}

\textbf{Phase in Motion Synthesis and Control.}  
Methods that represent motion in the frequency domain date back to the 1990s, including early work on motion synthesis~\cite{liu1994hierarchical} and editing~\cite{bruderlin1995motion}.  
With the advent of machine learning, phase-based representations have been widely adopted to improve temporal alignment and controllability in motion models.  
The \emph{Phase-Functioned Neural Network (PFNN)}~\cite{holden2017phase} introduced a phase-conditioned architecture for locomotion control, where network weights are modulated by a scalar phase inferred from foot contacts.  
Subsequent extensions~\cite{starke2019neural, starke2020local} generalized this idea to multi-contact and limb-specific phase conditioning.  
\textit{DeepPhase}~\cite{starke2022deepphase} further extended the concept to unstructured motion data by learning multi-dimensional phase variables, forming a phase manifold that captures complex periodic relationships.  
Follow-up works explored phase-based in-betweening~\cite{starke2023motion} and group choreography modeling~\cite{le2024group_coreo}.  
Recent approaches such as \textit{WalkTheDog}~\cite{li2024walkthedog} apply phase-space vector quantization with a shared motion vocabulary across species.  
Despite their effectiveness, these methods remain task-specific, tied to fixed skeletons, and not probabilistic, not allowing generative motion synthesis.  
Our approach directly overcomes these limitations by introducing a functional phase manifold that is both generative and skeleton-agnostic, while retaining the interpretability of phase-based representations.

\textbf{Motion Generative Models.}  
While a few studies have explored generation in phase space, such as \textit{PhaseDiff}~\cite{wan2023diffusionphase} and \textit{Compositional Phase Diffusion}~\cite{au2025transphase}, most approaches transfer existing generative paradigms directly to motion without domain-specific adaptations.  
Notable examples include diffusion-based models such as \emph{MDM} \cite{tevet2022human} and latent diffusion models such as \emph{MLD}~\cite{chen2023executing}, as well as autoregressive approaches like \emph{MotionGPT} \cite{jiang2023motiongpt}.  
A wide range of follow-up works have focused on scaling to larger datasets and models~\cite{wangscaling, fan2025go}, yet common issues persist, including implausible foot contacts and limited controllability.  
PhaseDiff operates in a predefined frequency domain, encoding motion into discrete periodic parameters using fixed frequency sets, without modifying the DeepPhase autoencoder.
\emph{Deep Compositional Phase Diffusion} \cite{au2025transphase} proposes a diffusion framework for long compositional motion generation, tokenizing phase variables over variable-length intervals to preserve semantic consistency across composed segments; like PhaseDiff, it builds on top of the original DeepPhase autoencoder rather than rethinking the underlying representation.
In contrast, FunPhase is a novel autoencoder that learns a continuous phase manifold and reconstructs motion as a spatio-temporal function. This enables sampling at arbitrary temporal resolutions and demonstrates superior expressibility, as evidenced in Table \ref{tab:recon}. Whereas these prior works target the generative side of the pipeline while inheriting a fixed phase encoder, our contribution is on the representation side: a functional phase manifold with continuous-time decoding, a different latent parameterization (including our phase transformation), and skeleton-agnostic modeling, enabling applications such as temporal super-resolution and partial-body completion. The two directions are complementary — our formulation is compatible with variable-length intervals and could be combined with Deep Compositional Phase Diffusion for long-horizon compositional generation.
Controllable human motion generation has been further explored in \emph{CAMDM}~\cite{chen2024taming}, which employs motion diffusion probabilistic models to produce diverse character animations that respond in real time to dynamic user control signals. 
Efforts toward generalization across skeletons include \emph{AnyTop}~\cite{gat2025anytop}, \emph{AniMo}~\cite{wang2025animo}, \emph{UniMoGen}~\cite{khani2025unimogen}, and \emph{SinMDM}~\cite{raab2023single}.  
\emph{SinMDM} learns the internal motifs of a single motion sequence with arbitrary topology, enabling long and diverse animations across humans, animals, and imaginary creatures.  
Most recently, \emph{ACMDM}~\cite{meng2025absolute} showed that representing motion with absolute joint coordinates can achieve state-of-the-art fidelity and diversity.  

Comprehensive surveys~\cite{zhu2023human, khani2025motion} review these advances, emphasizing the growing need for interpretable and temporally structured motion priors.  
Notably, all the above are \emph{frame-based} approaches, whereas our model reconstructs motion as a \emph{spatio-temporal function}, allowing variable-rate decoding and improved stability. Moreover, we show that incorporating phase features provides an effective inductive bias that regularizes the generative process and drastically improves denoising stability.

\textbf{Generative Models over Function Space.} 
Early attempts at generating function spaces~\cite{franzese2023continuous} demonstrated the potential of representing data as continuous mappings, including applications in domains such as biological processes~\cite{wang2023swallowing}.  
More recently, \textit{FunDiff}~\cite{wang2025fundiff} formalized diffusion over function space, enabling smooth and resolution-independent generative modeling.  
Their formulation provides natural regularization and compatibility with physical priors, properties that are particularly useful for motion generation.  
Our work draws inspiration from this functional perspective: FunPhase extends the concept to motion data, combining phase-manifold learning with a functional decoder for smooth, skeleton-agnostic motion synthesis.

\begin{figure*}[t]
    \centering
    \includegraphics[width=1\linewidth]{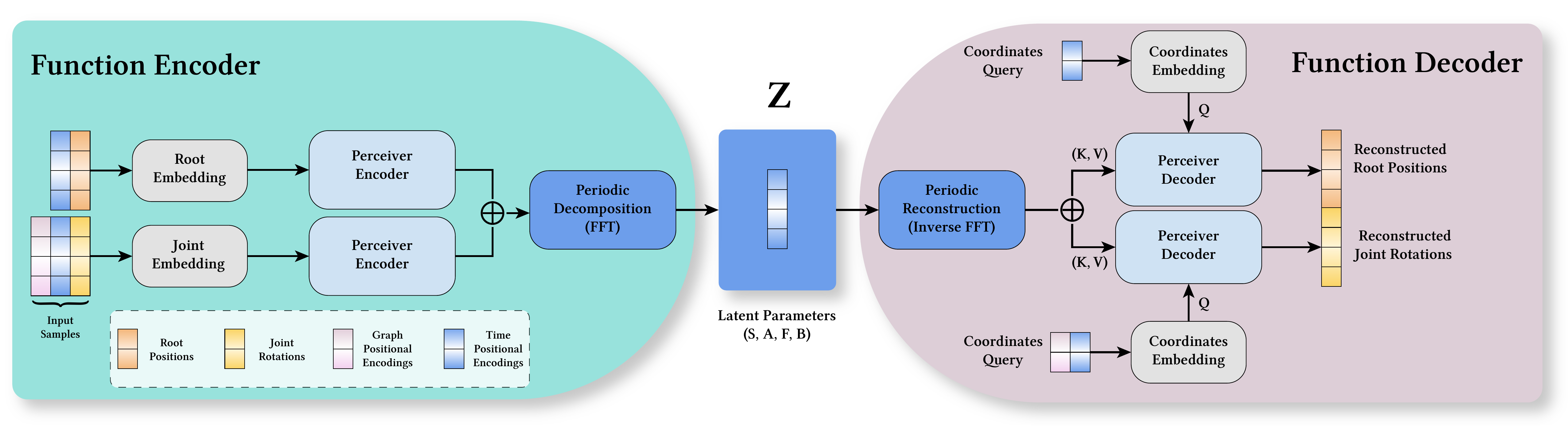}
    \caption{\textbf{Overview of the Periodic Function Autoencoder (FunPhase) architecture.} The figure illustrates the separated processing of joint rotations and root positions through Perceiver-based encoder–decoder modules. The latent space is decomposed by a Fast Fourier Transform (FFT) layer in its periodic components (Phase shift, Amplitude, Frequency, Bias) to achieve an even more compact representation and enforce periodicity. The latent space is then reconstructed with the inverse FFT, and the functions are evaluated at the coordinates given as input to the decoder.}
    \label{fig:model_arch}
\end{figure*}

\section{Background}
\label{sec:background}

The use of phase variables to describe motion progression has been well established in data-driven motion modeling, most notably in the \textit{Phase-Functioned Neural Network} (PFNN)~\cite{holden2017phase} and its extensions to multi-contact and limb-specific phase representations~\cite{starke2020local,mason2022_100style}. These methods rely on phase signals, often derived from contact heuristics, to align frames across motion sequences. Building upon this idea, \textit{DeepPhase}~\cite{starke2022deepphase} introduced the concept of a \emph{motion phase manifold}, which generalizes phase modeling to unstructured motion data.

\subsection{The Motion Phase Manifold}

DeepPhase proposed to learn multi-dimensional phase variables directly from motion data through an encoder-based latent space endowed with a frequency-domain inductive bias. The so-called \emph{Periodic Autoencoder (PAE)} models the temporal structure of motion as a combination of learned periodic components.  

Given an input motion sequence $\mathbf{X} \in \mathbb{R}^{D \times N}$, where $D$ is the number of motion features and $N$ is the number of frames, the encoder $\mathit{g}$ maps the sequence into a latent representation $L = \mathit{g}(\mathbf{X}) \in \mathbb{R}^{M \times N}$ with $M$ latent channels using $1D$ convolution. Each latent channel $\hat{l_c}$ is then parameterized as a sinusoidal function:
\begin{equation}\label{eq:latent-sine}
\hat{l_c} = a_c \cdot \sin(2\pi(f_c \cdot \mathcal{T} - s_c)) + b_c,
\end{equation}

where $a_c, f_c, s_c, b_c$ denote amplitude, frequency, phase shift, and offset, respectively, and $\mathcal{T}$ is the time window. The parameters $a_c, f_c, b_c$ are obtained via a differentiable real-valued Fast Fourier Transform (FFT) layer, while $s_c$ is predicted through a learned phase regressor.
This formulation enforces each latent channel to capture locally periodic motion components, such as gait cycles, arm swings, or torso oscillations. Finally, the decoder reconstructs the original motion through $1D$ deconvolutions $\mathbf{Y}=\mathit{h}([\hat{l_1},...,\hat{l_M}])$, minimizing the mean-squared error loss between input and output. %

From the periodic parameters, a \emph{phase manifold} $\boldsymbol{\mathcal{P}}(t) \in \mathbb{R}^{2M}$ is constructed:
\begin{align}\label{eq:manifold}    
\boldsymbol{\mathcal{P}}_{2i-1}^{(t)} &= a_i^{(t)} \cdot \sin(2\pi \cdot s_i^{(t)}), \\ \boldsymbol{\mathcal{P}}_{2i}^{(t)} &= a_i^{(t)} \cdot \cos(2\pi \cdot s_i^{(t)}).
\end{align}

This hyperspherical transformation couples amplitude and phase while discarding quasi-static parameters such as frequency and offset.  %

\subsection{Interpretation}

As discussed in~\cite{starke2022deepphase}, this formulation promotes clustering of motions both in space and time, yielding a smooth and interpretable manifold where temporal alignment emerges naturally. The learned phase manifold $\boldsymbol{\mathcal{P}}^{(t)}$ captures the rhythmic structure of motion across multiple body parts, forming smooth cyclic trajectories in low-dimensional projections (\autoref{fig:phases}). This representation facilitates downstream tasks such as motion matching, style transfer, and phase-based motion synthesis.  

However, the original DeepPhase framework is inherently limited by its \emph{frame-based convolutional} design and dependency on a \emph{fixed skeleton topology}, which constrains generalization across skeletons and datasets. %
Moreover, extending the model to probabilistic or generative settings, where one needs to sample from a learned motion distribution, is non-trivial.

In the following section, we build on these insights and introduce \textbf{FunPhase}, a functional extension of the Periodic Autoencoder. FunPhase reconstructs motion as a continuous spatio-temporal function rather than a discrete sequence, enabling scalable and skeleton-agnostic phase modeling suitable for generative motion synthesis.

\section{Methods}
\label{sec:method}

Our motion generation framework comprises two stages, conceptually similar to latent diffusion pipelines.
First, we introduce a Periodic Function Autoencoder (\textbf{FunPhase}) that learns a continuous periodic representation of movement in function space.
Our model maps discrete motion sequences to a compact latent space parameterized by periodic functions, enabling motion reconstruction and sampling at arbitrary temporal resolutions and facilitating downstream tasks such as motion synthesis and completion.
The architecture of FunPhase is summarized in \autoref{fig:model_arch} and described in detail in \autoref{sec:funphase-arch}.
Second, we train a diffusion model operating in the learned phase manifold, as detailed in \autoref{sec:diffusion-method}.  
This stage enables probabilistic motion synthesis in the same functional space, producing temporally coherent and physically plausible motion trajectories.

\subsection{FunPhase architecture}\label{sec:funphase-arch}

For our FunPhase design, we draw inspiration from the FunDiff architecture \cite{wang2025fundiff}. Our network is specifically adapted to model temporal sequences over a 3D skeletal graph.

\textbf{Motion representation.}
We represent joint rotations using the continuous $6D$ rotation representation \cite{zhou2018continuity}, which avoids the discontinuities and ambiguities inherent in other rotation parameterizations such as Euler angles or quaternions. 
For a skeleton with $J$ joints, the motion at frame $t$ is represented by a tensor of joint rotations $\mathbf{R}_t \in \mathbb{R}^{J \times 6}$.
The root joint position is treated separately from the rotational components to decouple global translation from local articulation. Root positions are represented in world coordinates as $\mathbf{X}_t^{root} \in \mathbb{R}^3$, encoding the 3D location of the pelvis or root joint at each frame.

\textbf{Positional Encoding.}
To provide the model with information about temporal structure and skeletal topology, we employ two types of positional encodings:

\textit{(i) Temporal Encoding:} temporal coordinates are mapped using Fourier features with a fixed range of frequencies.

\textit{(ii) Spatial Encoding:} skeletal structure is encoded using graph-based positional features derived from the Laplacian of the skeleton graph. For datasets with variable skeletal sizes, we instead use a heat-diffusion–based encoding that captures multi-scale structural information.

Full mathematical details are provided in the Appendix \ref{sec:implementation-details}.

\textbf{Encoder.}
We employ separate encoding pathways for joint rotations and root positions, each with dedicated Perceiver-based encoders \cite{jaegle2021perceiver}. This architectural choice captures the distinct dynamics of global translation and local articulation.
The \textit{joint} encoder processes the concatenation of joint rotations and spatial-temporal positional encodings
\(
\mathbf{h}_t^{joints} = [\mathbf{R}_t, \mathbf{p}^{graph}, \mathbf{p}_t^{time}]
\)
.
The Perceiver encoder consists of cross-attention blocks followed by self-attention blocks. It learns a set of $L_{joints}$ latent tokens $\mathbf{Z}^{joints} \in \mathbb{R}^{L_{joints} \times d_{joints}}$ that attend to the input sequence via cross-attention:
\[
\mathbf{Z}^{joints} = \text{PerceiverEnc}^{joints}(\mathbf{h}_{1:T}^{joints})
\]

Similarly, the \textit{root} encoder processes root positions concatenated with temporal encodings
\(
\mathbf{h}_t^{root} = [\mathbf{X}_t^{root}, \mathbf{p}_t^{time}]
\),
producing latent tokens $\mathbf{Z}^{root} \in \mathbb{R}^{L_{root} \times d_{root}}$.

The joint and root latents are independently projected to a common dimensionality $d_{latent}$ via linear layers, then concatenated and passed through a $1D$ convolutional bottleneck with circular padding to produce the unified latent representation:
\[
\mathbf{Z} = \text{Conv1D}([\mathbf{Z}^{joints}; \mathbf{Z}^{root}]) \in \mathbb{R}^{C \times d_{latent}}
\]
where $C$ is the number of latent channels.

\textbf{Phase Decomposition integration.}
The FunPhase framework enables channel-wise periodic decomposition following the DeepPhase approach \cite{starke2022deepphase}.
Following their Periodic Autoencoder, we use a combination of the Fast Fourier Transform and a linear layer to obtain the periodic parameters, modeling each latent channel as a sinusoid.

Each latent channel is parameterized by four periodic components:
\[
\boldsymbol{\theta}_c = [s_c, a_c, f_c, b_c] \in \mathbb{R}^4.
\]

This compact representation encodes the complete periodic structure of motion in $4C$ parameters, where $C$ ranges from $16$ to $256$ in our implementation, resulting in a compact representation with $\leq 1024$ parameters per clip.

\textbf{Decoder.}
The decoder reconstructs the latent function by evaluating the learned sinusoidal parameters (\autoref{eq:latent-sine}).
This reconstructed latent representation $\hat{\mathbf{Z}}$ is then passed through the inverse bottleneck convolution and split into joint and root latents:
\[
[\hat{\mathbf{Z}}^{joints}; \hat{\mathbf{Z}}^{root}] = \text{DeConv1D}(\hat{\mathbf{Z}})
\]

The joint and root decoders use cross-attention to query the reconstructed latents at arbitrary spatio-temporal positions. For a query time \(t'\) and joint  \(j'\), the decoder produces:
\[
\begin{aligned}
\hat{\mathbf{R}}_{t',j'} &= \text{PerceiverDec}^{joints}(\hat{\mathbf{Z}}^{joints}, \mathbf{p}_{j'}^{graph}, \mathbf{p}_{t'}^{time}) \\
\hat{\mathbf{X}}_{t'}^{root} &= \text{PerceiverDec}^{root}(\hat{\mathbf{Z}}^{root}, \mathbf{p}_{t'}^{time})
\end{aligned}
\]

This allows the model to sample motion at any temporal resolution, enabling applications like temporal super-resolution and motion retiming, and at any joint, enabling body completion.

\textbf{Training Objective.}
The model is trained to minimize a combination of reconstruction losses:

\textit{(i) Rotation Loss:} Geodesic distance for joint rotations:
\begin{equation}\label{eq:geodesic}
    \mathcal{L}_{rot} = \frac{1}{TJ}\sum_{t,j} \arccos\left(\frac{\text{tr}(\mathbf{R}_{t,j}\hat{\mathbf{R}}_{t,j}^{\top}) - 1}{2}\right)  
\end{equation}

\textit{(ii) Root Position Loss:} Mean squared error for root positions:
\begin{equation}
    \mathcal{L}_{root} = \|\mathbf{X}^{root} - \hat{\mathbf{X}}^{root}\|_2^2    
\end{equation}

\textit{(iii) Forward Kinematics Loss:} To enforce physical plausibility, we penalize deviations in forward-kinematics (FW) joint positions:
\begin{equation}
  \mathcal{L}_{FK} = \|\text{FK}(\mathbf{R}, \mathbf{X}^{root}) - \text{FK}(\hat{\mathbf{R}}, \hat{\mathbf{X}}^{root})\|_2^2  
\end{equation}
together with foot penetration (FP) and foot sliding (FS) penalties:
\begin{equation}
\begin{split}
\mathcal{L}_{foot} = {} & \|\text{FP}(\mathbf{R}, \mathbf{X}^{root}) - \text{FP}(\hat{\mathbf{R}}, \hat{\mathbf{X}}^{root})\|_2^2 \\
& + \|\text{FS}(\mathbf{R}, \mathbf{X}^{root}) - \text{FS}(\hat{\mathbf{R}}, \hat{\mathbf{X}}^{root})\|_2^2
\end{split}
\end{equation}

The total loss is:
\begin{equation}\label{eq:total-loss}
    \mathcal{L} = 0.5 (\mathcal{L}_{rot} + \mathcal{L}_{root}) + 0.5 (\mathcal{L}_{FK} + 0.01 \mathcal{L}_{foot})
\end{equation}

The effectiveness of this composite loss and other design choices is validated in the Ablation section of the Supplementary materials. %

\subsection{Phase diffusion}\label{sec:diffusion-method}

Building upon the FunPhase autoencoder, we introduce a latent diffusion model that operates directly on the periodic function parameters, enabling class-conditional motion generation in a compact, semantically meaningful space.

\textbf{Phase Transformation.}
The periodic parameterization \(\boldsymbol{\theta}_c = [s_c, a_c, f_c, b_c]\) from FunPhase, while compact and interpretable, poses challenges for diffusion modeling due to the domain and distribution of some parameters. We apply domain transformations to ensure compatibility with Gaussian diffusion.
The phase \(s_c \in [0, 1)\) is transformed to Cartesian coordinates to handle periodicity:
\[
\begin{aligned}
\mathbf{a}_{c}^{\cos} &= a_c \cos(2\pi s_c), \quad \quad
\mathbf{a}_{c}^{\sin} = a_c \sin(2\pi s_c).
\end{aligned}
\]
This representation avoids discontinuities at the phase boundary and encodes both amplitude and phase information jointly.

The frequency \(f_c \in [0, f_{max}]\) is unbounded via probit transformation:
\[
f_c^{probit} = \sqrt{2} \cdot \text{erf}^{-1}\left(2 \cdot \frac{f_c}{f_{max}} - 1\right)
\]
where \(f_{max} = 0.5 \cdot d_{latent} / (2\pi)\) and erf stands for the error function. This maps the bounded frequency domain to the entire real line \(\mathbb{R}\), making it suitable for Gaussian noise injection.

The offset \(b_c\) is unbounded and requires no transformation. The final diffusion-compatible representation is:
\begin{equation}\label{eq:phase-transform}
    \boldsymbol{\theta}_c^{diff} = [\mathbf{a}_{c}^{\cos}, \mathbf{a}_{c}^{\sin}, f_c^{probit}, b_c] \in \mathbb{R}^4
\end{equation}

After sampling, we recover the original periodic parameters:
\[
\begin{aligned}
a_c &= \sqrt{(\mathbf{a}_{c}^{\cos})^2 + (\mathbf{a}_{c}^{\sin})^2 + \epsilon} \\
s_c &= \frac{1}{2\pi} \arctan2(\mathbf{a}_{c}^{\sin}, \mathbf{a}_{c}^{\cos}) \\
f_c &= f_{max} \cdot \frac{1}{2}\left(1 + \text{erf}\left(\frac{f_c^{probit}}{\sqrt{2}}\right)\right)
\end{aligned}
\]

In the Appendix \ref{app:phaseTransf}, we provide a further analysis on the advantage of this choice.

\textbf{Diffusion Model.}
We employ a Diffusion Transformer (DiT) \cite{peebles2023scalable} architecture adapted for 1D latent sequences. 
We use the \emph{velocity parameterization}, in which the network predicts the instantaneous velocity field of the reverse diffusion process. This formulation provides more stable training and sampling compared to standard noise prediction \cite{zheng2023improved}. At each timestep, the model outputs a velocity vector integrated with a linear noise schedule to obtain the latent trajectory. After sampling, we invert the latent transformations and decode motion using FunPhase.
The model is conditioned on class labels and partial motion inputs: known keyframes or joint rotations are encoded with the pretrained FunPhase encoder, and the resulting latents are concatenated with the class embedding.
Further architectural and diffusion details are provided in the Appendix \ref{sec:implementation-details}.

\begin{table*}[t]
\centering
\caption{\textbf{Autoencoder comparison on the \textsc{Dog} and \textsc{100Style} datasets.}
FunPhase consistently outperforms DeepPhase in all metrics, showing more accurate and physically consistent reconstructions.}
\small
\begin{tabular}{llcccccc}
\toprule
Dataset & Method & Position (cm) $\downarrow$ & Orientation $\downarrow$ & NPSS 
$\downarrow$ & Sliding $\downarrow$ & Penetration $\downarrow$ & ACL $\downarrow$ \\
\midrule\midrule
\multirow{2}{*}{\textsc{Dog}} & DeepPhase {\footnotesize-16C}  &               144 &       0.54 &  3.58 &   1.76 &  0.912 & 
1.768 \\
\phantom{Dataset} & \textbf{FunPhase} {\footnotesize-16C} & \textbf{61.4} & \textbf{0.34} & \textbf{1.83} & \textbf{0.21} & \textbf{0.351} & \textbf{1.062} \\
\midrule
\multirow{6}{*}{\textsc{100Styles}} & DeepPhase{\footnotesize-32C} & 92.9 & 0.35 & 3.68 & 1.47 & 0.414 & 1.504 \\
\phantom{Dataset} & MLD{\footnotesize-VAE} & 59.8 & 0.29 & 2.93 & 0.88 & 0.211 & 1.380 \\
 \phantom{Dataset} & ACMDM{\footnotesize-AE} & \textbf{0.32} & \textbf{0.01} & \textbf{0.28} & 0.33 & 0.003 & 1.412 \\
\phantom{Dataset} & Function AE{\footnotesize-256C} & 0.74 & \underline{0.05} & \underline{0.59} & 0.23 & 0.001 & 1.383 \\
\phantom{Dataset} & \textbf{FunPhase}{\footnotesize-32C} & 1.93 & 0.10 & 1.18 & \textbf{0.16} & \underline{0.001} & \textbf{1.371} \\
\phantom{Dataset} & \textbf{FunPhase}{\footnotesize-256C} & \underline{0.36} & 0.20 & 0.75 & \underline{0.21} & \textbf{0.001} & \underline{1.378} \\
\bottomrule
\end{tabular}
\label{tab:recon}
\end{table*}

\section{Experiments}\label{sec:results}

We evaluate our method on reconstruction (\ref{sec:pfae-res}), latent diffusion generation (\ref{sec:diffusion-res}), and motion prediction  (\ref{sec:controller}), and report ablations in Appendix~\ref{app:ablation}.

\textbf{Datasets.}
To assess robustness across characters and motion types, we evaluate on both human and animal motion datasets.
We use \textsc{100Style}~\cite{mason2022_100style} for stylized human locomotion, \textsc{Dog}~\cite{zhang2018mode} for quadruped motion with frequent mode transitions, and \textsc{Truebones ZOO}~\cite{truebones} for large-scale multi-skeleton evaluation.
Unless otherwise specified, sequences are divided into fixed-length 60-frame windows.
Additional preprocessing details are provided in the Appendix \ref{app:datasets}.

\textbf{Metrics.}
We evaluate reconstruction and generation quality using standard metrics for spatial accuracy, temporal coherence, physical plausibility, and perceptual fidelity.
These include joint position and rotation errors, NPSS, physics-based measures such as foot sliding and acceleration smoothness, as well as FID, classification accuracy, and diversity for generation.
For \textsc{Truebones ZOO}, we additionally report coverage and diversity metrics following~\cite{gat2025anytop}.
Complete metric definitions and implementation details are provided in Appendix~\ref{app:metrics}.

\textbf{Baselines.} 
We first compare our FunPhase model with DeepPhase~\cite{starke2022deepphase} on the reconstruction task using both \textsc{100Style} and \textsc{Dog} datasets, and provide visual comparisons for the motion controller.
In the generative setting, to our knowledge, our work is the first to explore motion generation in a \emph{function space}.
Nevertheless, we compare with state-of-the-art latent diffusion approaches: the Absolute Coordinate Motion Diffusion Model (ACMDM) \cite{meng2025absolute}, the original Latent Motion Diffusion model (LDM) \cite{chen2023executing}, and the Conditional Autoregressive Motion Diffusion Model (CAMDM) \cite{chen2024taming}, which achieves state-of-the-art performance on stylized human locomotion (\textsc{100Style}).
To further validate our autoencoder, we also compare the autoencoder component of ACMDM and the VAE component of LDM against our FunPhase autoencoder.
For multi-skeleton settings, we further compare with AnyTop \cite{gat2025anytop} and SinMDM \cite{raab2023single}.
We use the released checkpoints for CAMDM, AnyTop, and retrain ACMDM and LDM on the \textsc{100Style} dataset using their official code and settings.
Finally, we include our Function Autoencoder without the phase decomposition as an additional baseline for both reconstruction and generation tasks.

\begin{table*}[ht]
    \centering
    \caption{\textbf{Latent diffusion results on the \textsc{100Style} dataset.}
Comparison between FunPhase and state-of-the-art motion latent diffusion baselines.
FID and Accuracy evaluate perceptual fidelity and condition alignment, respectively.
Diversity measures pose variation across generated samples, while Foot Sliding, Coherence, and ACCL assess physical realism.}
    \begin{tabular}{lcccccc}
    \toprule
    Method & FID $\downarrow$ & Accuracy (\%) $\uparrow$ & Diversity $\uparrow$ & Foot Sliding $\downarrow$ & Coherence $\uparrow$ & ACL $\downarrow$ \\ 
    \midrule\midrule
    MLD & $1.99\textsuperscript{\scriptsize $\pm$1.22}$ & 41.88 & $0.65\textsuperscript{\scriptsize $\pm$0.01}$ & 1.16 & 1.06 & 1.34 \\
    ACMDM & $5.45\textsuperscript{\scriptsize $\pm$3.04}$ & 15.67 & $\mathbf{0.75}\textsuperscript{\scriptsize $\pm$0.01}$ & 1.74 & \textbf{1.53} & 2.46 \\
    CAMDM & $\underline{0.91}\textsuperscript{\scriptsize $\pm$0.62}$ & \textbf{88.22} & $\underline{0.69}\textsuperscript{\scriptsize $\pm$0.01}$ & \underline{0.69} & 1.08 & 6.02 \\
    Function Diff. & $1.19\textsuperscript{\scriptsize $\pm$0.43}$ & 34.27 & $0.54\textsuperscript{\scriptsize $\pm$0.01}$ & 1.23 & 1.14 & \textbf{1.25} \\
    \textbf{FunPhase} & $\textbf{0.51}\textsuperscript{\scriptsize $\pm$0.16}$ & \underline{76.17} & $0.64\textsuperscript{\scriptsize $\pm$0.01}$ & \textbf{0.52} & \underline{1.17} & \underline{1.33} \\
    \bottomrule
    \end{tabular}
    \label{tab:diffusion}
\end{table*}

\subsection{FunPhase}\label{sec:pfae-res}
In this section, we assess the quality of the learned Phase Manifold and motion function in comparison with baseline models. We show that the Phase Manifold learned by our approach preserves the core behavior of DeepPhase while significantly improving the expressiveness and fidelity of motion reconstruction.
We visualize the learned phase manifold and compare it with DeepPhase and raw motion features in Appendix \ref{app:phaseManifold}, showing that our representation preserves the circular structure characteristic of cyclic motion.

\textbf{Reconstruction.}
To demonstrate that our model can accurately infer motion functions, we evaluate reconstruction error and physical plausibility of reconstructed motions, and compare FunPhase against baseline methods in \autoref{tab:recon}.
FunPhase substantially improves over DeepPhase in its ability to extract phase-aware motion representations, achieving lower reconstruction errors and consistently better physics-based scores.
The higher absolute errors on \textsc{Dog} are likely due to the dataset’s limited size, 
as well as the presence of frequent transitions between distinct periodic behaviors, which challenge purely phase-based representations. Despite these challenges, FunPhase consistently outperforms DeepPhase across all metrics, indicating greater robustness to mixed-period motions. In Particular, FunPhase reduces position error by $\approx57\%$ (144 → 61.4 cm) and NPSS by $\approx49\%$, while also substantially lowering sliding (1.76 → 0.21) and penetration (0.912 → 0.351).  
On \textsc{100Style}, while ACMDM achieves slightly lower reconstruction errors, indicating strong frame-wise fidelity, this does not consistently translate into physically plausible motion, as evidenced by higher sliding and ACL scores compared to FunPhase.  In contrast, FunPhase trades a small amount of raw reconstruction accuracy for markedly improved physical consistency, reflecting the regularizing effect of phase-based motion decomposition.
Increasing latent capacity (32C → 256C) improves reconstruction accuracy while largely preserving physical consistency, suggesting that FunPhase scales gracefully with model capacity and does not rely on over-compression to achieve plausible motion.
Despite using a substantially more compact latent space, FunPhase outperforms the Function Autoencoder in most metrics, including position error and sliding, while remaining competitive in orientation and NPSS. This indicates that introducing a periodic component in the latent space does not compromise representational power; instead, it enables a more efficient and structured encoding of motion with improved performance.
Moreover, the functional and skeleton-agnostic formulation of FunPhase enables reconstruction from partial skeletons and incomplete temporal windows, as demonstrated in the temporal super-resolution examples in \autoref{fig:recon} and the controlled ablations in Appendix \ref{app:increasing_dist}.
Overall, these results show that FunPhase achieves a favorable balance between reconstruction accuracy and physical plausibility, outperforming prior phase-based methods while remaining competitive with state-of-the-art autoencoders that do not explicitly model periodic structure and that are not skeleton-agnostic.

\begin{figure}
    \centering
    \includegraphics[width=1\linewidth]{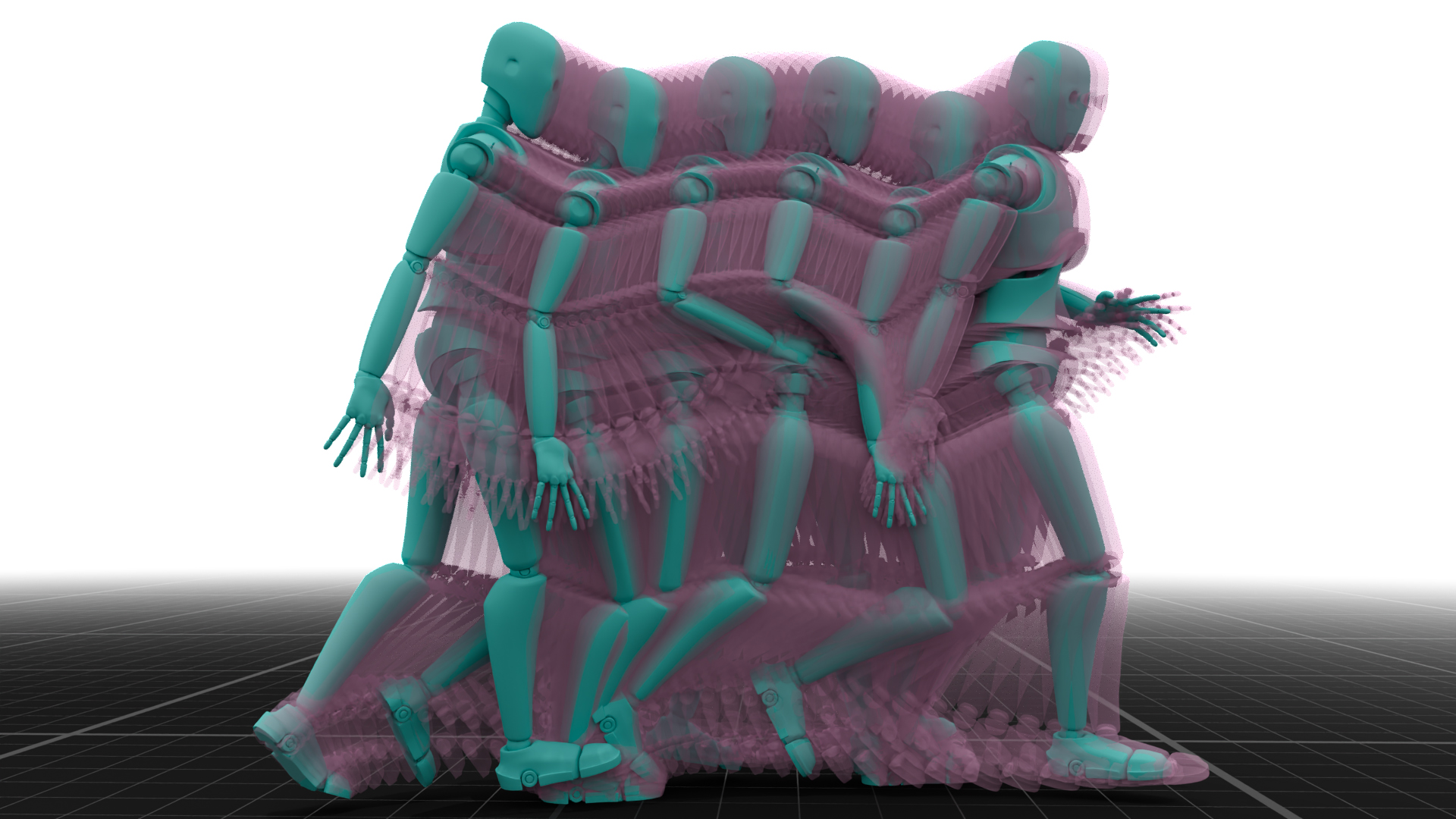}
    \caption{\textbf{FunPhase super-resolution.} Given a sparse set of keyframes, FunPhase reconstructs the full continuous motion while preserving physical plausibility.}
    \label{fig:recon}
\end{figure}

\subsection{Latent diffusion in Phase Manifold}\label{sec:diffusion-res}

We perform diffusion directly in the phase manifold as described in \ref{sec:diffusion-method}. Quantitative results are reported in \autoref{tab:diffusion} for the human dataset \textsc{100Style} and in \autoref{tab:zoo} for the animal dataset \textsc{Zoo}. In the human case, the class conditioning is on locomotion style, while for animals on the species.

\begin{figure}
    \centering
    \includegraphics[width=1\linewidth]{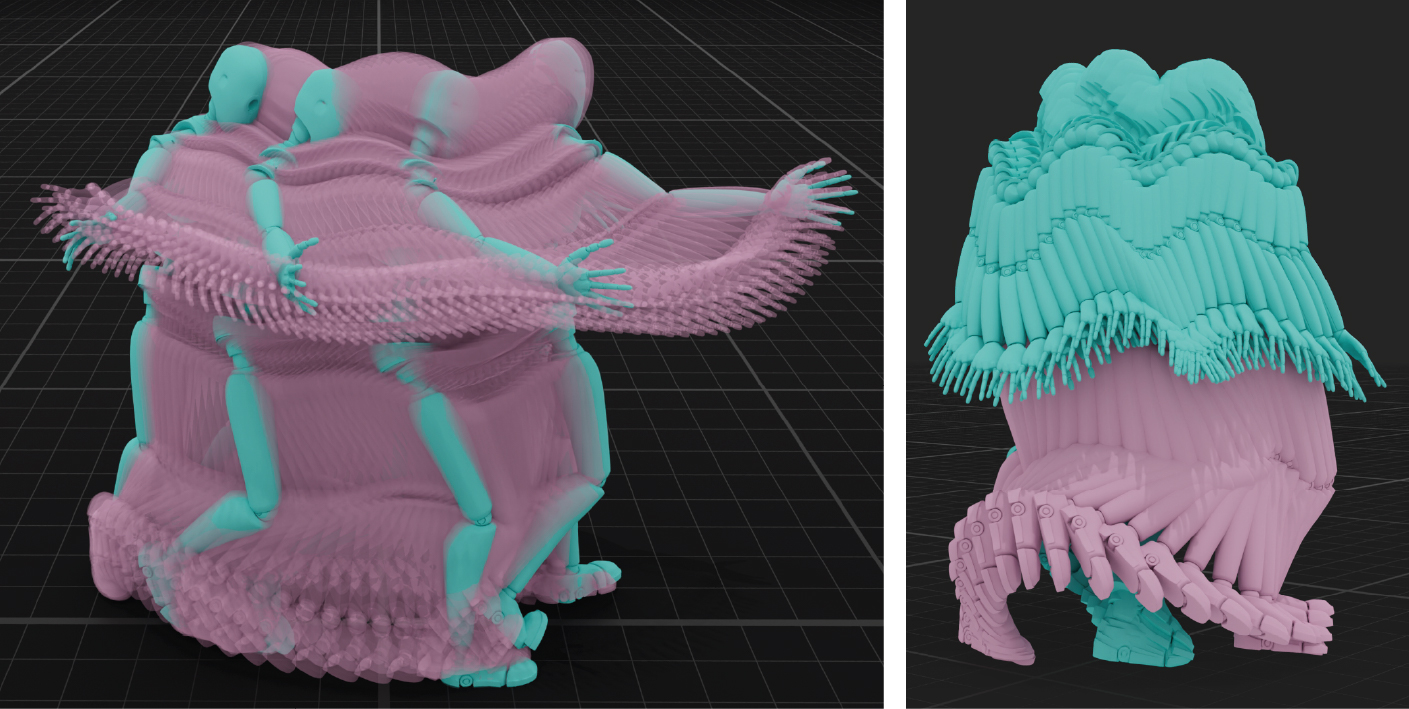}
    \caption{\textbf{Diffusion examples on \textsc{100STYLE}.} On the left we show and example of generation from a sparse set of key frames (in green). On the right we show an example of body completion of the right leg (in pink).} 
    \label{fig:diffusion}
\end{figure}

Among all methods, FunPhase achieves the lowest FID and foot-sliding error on \textsc{100Style}, indicating both high visual fidelity and superior physical realism. While CAMDM attains higher conditioning accuracy, it exhibits substantially larger acceleration discontinuities (ACL), suggesting reduced physical plausibility. In contrast, FunPhase maintains competitive accuracy while significantly improving physical consistency, resulting in a more favorable overall trade-off.
 Compared to Function Diffusion, which also operates in a continuous functional space, FunPhase leverages an explicit phase-based periodic component in the latent space, which improves denoising stability and generation quality.
On the Zoo dataset, FunPhase achieves the highest coverage and the lowest intra-class diversity discrepancy, indicating more consistent generation across species while avoiding mode collapse. Although AnyTop attains higher raw diversity, this comes at the cost of reduced coverage and stability.
Beyond quantitative gains, our functional formulation enables generation under partial conditioning, such as from sparse keyframes or incomplete body observations. Examples of such scenarios are shown in \autoref{fig:diffusion}.

\begin{table}[th]
    \small
    \centering
    \caption{\textbf{Latent diffusion results on the \textsc{Zoo} dataset.}
We compare our method with other multi-skeleton baselines.}
    \begin{tabular}{lcccc}
    \toprule
    \multirow{2}{*}[0.4em]{Method} & 
    \multirow{2}{*}[0.4em]{Cov. $\uparrow$} & 
    \makecell{Div. $\uparrow$ \\(Local)} & 
    \makecell{Div. $\uparrow$ \\(Inter)} & 
    \makecell{Intra Div. $\downarrow$ \\Diff.} \\
    \midrule\midrule
    SMDM & $\underline{89}\textsuperscript{\scriptsize $\pm$ 15}$ &  $0.08\textsuperscript{\scriptsize $\pm$0.13}$ &  $\underline{0.28}\textsuperscript{\scriptsize $\pm$0.13}$  &  $0.14\textsuperscript{\scriptsize $\pm$0.09}$\\
    AnyTop & $80\textsuperscript{\scriptsize $\pm$22}$ &  $\textbf{0.26}\textsuperscript{\scriptsize $\pm$0.12}$  &  $\textbf{0.37}\textsuperscript{\scriptsize $\pm$0.18}$  &  $\underline{0.14}\textsuperscript{\scriptsize $\pm$0.08}$ \\
    \textbf{FunPhase} & $\textbf{96}\textsuperscript{\scriptsize $\pm$7}$ & $\underline{0.11}\textsuperscript{\scriptsize $\pm$0.04}$  &  $0.21\textsuperscript{\scriptsize $\pm$0.05}$  &  $\textbf{0.06}\textsuperscript{\scriptsize $\pm$0.03}$  \\
    \bottomrule
    \end{tabular}
    \label{tab:zoo}
\end{table}

\subsection{Motion Prediction}\label{sec:controller}

To further validate the learned phase manifold, we train a Neural Motion Controller within our phase space following the setup of \cite{starke2022deepphase}. The controller adopts a Weight-Blended Mixture-of-Experts architecture similar to \cite{starke2020local, zhang2018mode}, but instead of relying on velocities or contact-based local phases as input to the gating network, it uses the phase vectors from our learned manifold (\autoref{eq:manifold}). This design enables the model to generate motion in an autoregressive manner while operating in a more expressive space than DeepPhase.
The controller produces coherent motion sequences conditioned on both style and user inputs such as trajectories. Qualitative examples and a visual comparison with DeepPhase are presented in \autoref{fig:controller} and in the supplemental video.

\begin{figure}
    \centering
    \includegraphics[width=1\linewidth]{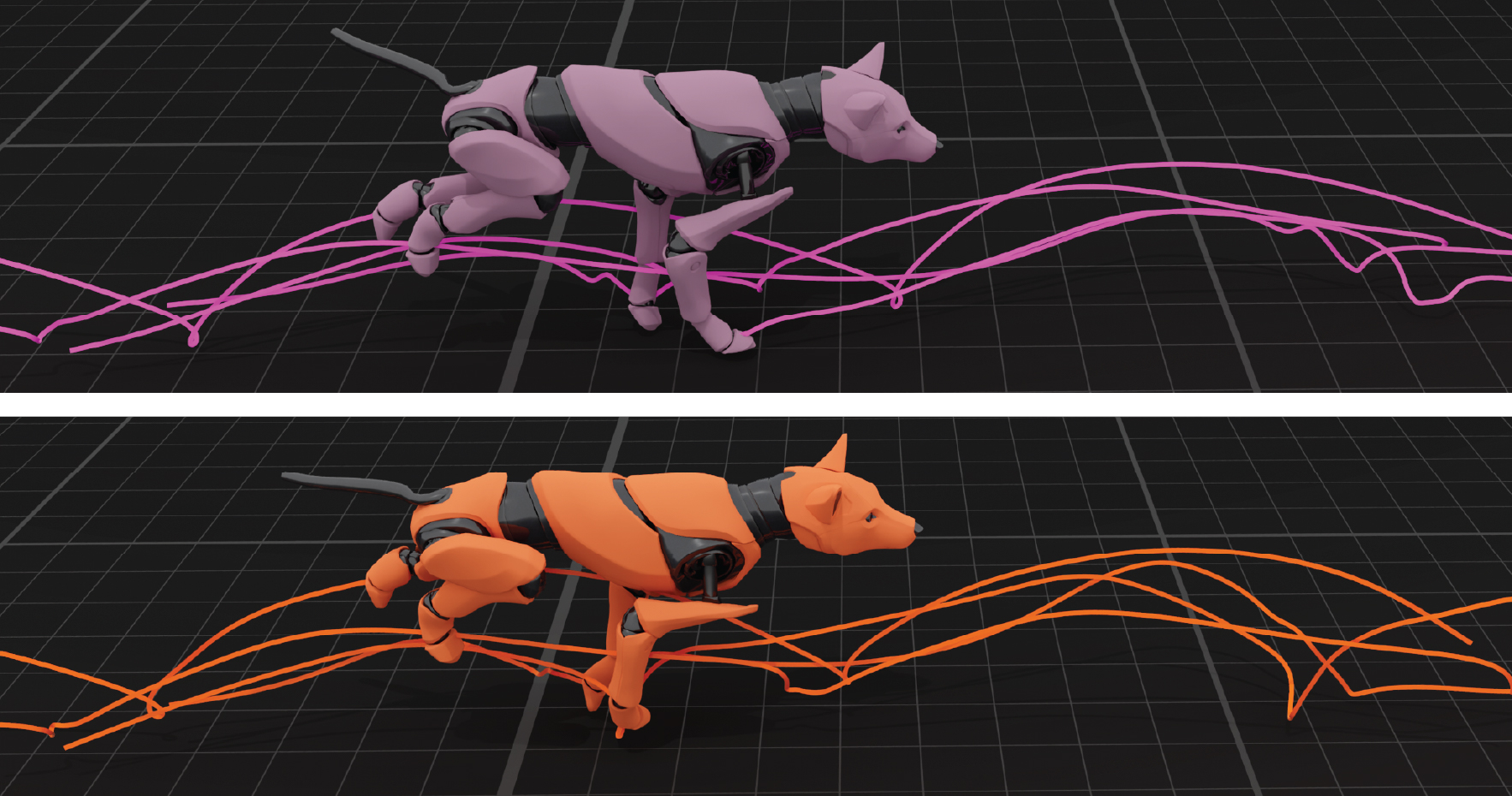}
    \caption{\textbf{Motion controller generation.} The top row shows the motion controller trained on FunPhase’s phase space, and the bottom row shows the controller trained on DeepPhase’s phase space. FunPhase enables the generation of smooth and realistic movements comparable to those produced by DeepPhase, highlighting the advantages of the phase manifold.}
    \label{fig:controller}
\end{figure}

\section{Conclusion and Future Work}

We introduced FunPhase, a function-space autoencoder for motion that represents movement as a continuous spatio-temporal function, with periodic decomposition of the latent space serving as regularizer. The function-space formulation provides skeleton-agnostic encoding, arbitrary-resolution decoding, and a unified framework for prediction, completion, and generation; the phase regularization, in turn, improves physical plausibility in reconstruction and denoising stability in diffusion-based generation.
While the applications presented in this work primarily serve as proof-of-concept demonstrations, the overall capabilities of our approach (both in terms of superior quantitative performance and flexibility) highlight its strong potential. In particular, the finding that incorporating phase decomposition into a functional autoencoder consistently improves motion generation quality is a key result that opens up new directions for modeling structured temporal dynamics.

At the same time, the reliance on phase-based representations introduces intrinsic limitations when modeling weakly periodic or fundamentally non-periodic motions, such as abrupt transitions. In such cases, enforcing a global phase structure may oversmooth temporal dynamics or limit expressiveness. Addressing these limitations is therefore a promising direction for future work.
A further limitation concerns the scope of our evaluation, which focuses  on locomotion-dominant datasets (100\textsc{Style}) and class-conditional generation. Extending  FunPhase to text-to-motion benchmarks such as HumanML3D and KIT-ML, where motions are semantically diverse and often only weakly periodic,  is an important direction for future work.
One avenue is to combine phase-based components with non-periodic latent variables or locally adaptive transformations. Additionally, incorporating physics-based priors could further improve realism and generalization in complex motion scenarios. Finally, extending the framework to applications such as motion in-betweening, partial-body completion, and long-horizon motion synthesis would further broaden the practical impact of functional phase representations.

\section*{Impact Statement}
This paper presents a method for learning structured, interpretable representations for motion modeling, with the goal of advancing research in machine learning for motion generation and prediction. The proposed approach may benefit applications in animation, simulation, and robotics by enabling more physically plausible and controllable motion synthesis. We do not anticipate significant negative societal impacts arising directly from this work.

\bibliography{main}
\bibliographystyle{icml2026}

\newpage
\appendix
\onecolumn

\section{Implementation Details}
\label{sec:implementation-details}

This section provides additional implementation details and the experimental configurations. The code is available at \href{https://github.com/Marco-Peg/FunPhase}{https://github.com/Marco-Peg/FunPhase}

\subsection{FunPhase}

\textbf{Positional Encoding.}
To provide the model with information about temporal structure and skeletal topology, we employ two types of positional encodings:

\textit{(i) Temporal Encoding:} We use Fourier positional encoding for temporal coordinates. For frame time \(t\), we compute:
\(
\mathbf{p}_t^{time} = [\sin(2\pi f_1 t), \cos(2\pi f_1 t), \ldots, \sin(2\pi f_K t), \cos(2\pi f_K t)]
\)
where the frequencies \(f_k\) are geometrically spaced and $K$ is the maximum number of frequencies.

\textit{(ii) Spatial Encoding:} We encode skeletal topology using eigenvectors of the normalized graph Laplacian, providing information about joint connectivity and structural relationships. For a skeleton graph with adjacency matrix $\mathbf{A} \in \mathbb{R}^{J \times J}$ and degree matrix $\mathbf{D} \in \mathbb{R}^{J \times J}$, we compute the normalized Laplacian $\mathbf{L} = \mathbf{D}^{-1/2}(\mathbf{D} - \mathbf{A})\mathbf{D}^{-1/2}$ and use its first $N$ eigenvectors as positional encodings \(\mathbf{p}_j^{graph} \in \mathbb{R}^N\) for joint \(j\), excluding the first constant eigenvector.
In the case of the dataset composed of graphs with different sizes, we encode the skeletal graph structure using the graph heat diffusion operator \( \mathbf{H}_t = e^{-t\mathbf{L}} \) at time $t > 0$. 
Let $\mathcal{T} = \{ t_1, \dots, t_T \}$ be a set of heat diffusion times, and let 
$r^{(q)} \sim \mathcal{N}(0, I_J)$ for $q = 1, \dots, Q$ be random probe vectors.
For each pair $(t_i, q)$, we compute the diffused signal
\( p_{t_i}^{(q)} = \mathbf{H}_{t_i} r^{(q)} = e^{-t_i L} r^{(q)} \).
The resulting spatial encoding is constructed by concatenating all diffused probes:
\[
\mathbf{P}^{graph} = \big[\, p_{t_1}^{(1)}, \dots, p_{t_1}^{(Q)},\, p_{t_2}^{(1)}, \dots, p_{t_T}^{(Q)} \,\big]
    \in \mathbb{R}^{J \times N}.
\]
In our implementation we set $Q=3$ and take $T=7$ heat diffusion times in the logspace between $10^{-2}$ and $10^0$.

\textbf{Architecture.}
In the largest model variant, which we use for the \textsc{100STYLE} dataset \cite{mason2022_100style}, the architecture employs $256$ latent channels, with separate encoders and decoders for joints and root. 
The joint encoder maps inputs to $256$-dimensional embeddings and processes them through $5$ Perceiver encoder blocks, each with depth $1$.
The root encoder uses $64$ latents with $128$-dimensional embeddings, processed through $3$ encoder blocks with depth $1$. Both decoders use $5$ and $3$ blocks respectively with depth $1$. The model contains approximately $34M$ parameters.
We use $60$-frame motion clips ($1$ second at $60$ fps), which we randomly subsample along both temporal and graph dimensions.
At training time we use the AdamW optimizer with a learning rate of $1e-4$. A cosine learning-rate schedule with warm-up and gradient clipping at $0.5$ is applied. %

\subsection{Diffusion model}
Our latent diffusion model takes as input a noisy phase-latent vector and a set of conditioning signals: diffusion timestep, class label, and partial motion input.
The conditioning are embedded and concatenated in a single vector $\mathbf{c}$.

\textbf{Backbone.} The core architecture consists of:
\begin{itemize}
    \item Linear Embedding: Projects each latent token from dimension 4 to embedding dimension \(d_{embed} = 256\) to obtain a new latent representation $\mathbf{h}$.
    \item Positional Encoding: 1D sinusoidal positional embeddings provide positional information across the latent channels
    \item DiT Blocks: \(L=8\) transformer blocks with adaptive layer normalization (AdaLN) for conditioning
    \item Output Projection: Projects back to the latent dimension 4
\end{itemize}

Each DiT block employs adaptive layer normalization modulated by the conditioning signal:
\[
\text{AdaLN}(\mathbf{h}, \mathbf{c}) = \gamma(\mathbf{c}) \odot \text{LayerNorm}(\mathbf{h}) + \beta(\mathbf{c})
\]
where \(\gamma\) and \(\beta\) are scale and shift parameters predicted from the conditioning vector \(\mathbf{c}\).

\textbf{Conditioning Mechanism. }The model is conditioned on three signals:

\textit{(i) Timestep Embedding:} The diffusion timestep \(t\) is encoded via learned sinusoidal embeddings followed by an MLP:
\[
\mathbf{e}_t = \text{MLP}(\text{SinusoidalEmbed}(t)) \in \mathbb{R}^{256}.
\]

\textit{(ii) Class Embedding:} Motion class labels \(y \in \{1, \ldots, K\}\) are embedded via a learned embedding layer:
\[
\mathbf{e}_y = \text{Embed}(y) \in \mathbb{R}^{64}.
\]

\textit{(iii) Partial Motion:} Optionally, a subset of joint rotations and root positions can be provided as conditioning input. These are encoded through the FunPhase encoder, and the resulting latent vector is then embedded via a learned embedding layer:
\[
\mathbf{e}_{context} = \text{Embed}_{context} (D_{FunPhase}([\Tilde{\mathbf{R}};\Tilde{\mathbf{X}}^{root}]) \in \mathbb{R}^{256}.
\]

These embeddings are concatenated and projected to form the conditioning vector:
\[
\mathbf{c} = \text{MLP}([\mathbf{e}_t; \mathbf{e}_y; \mathbf{e}_{context}]) \in \mathbb{R}^{256}.
\]

\textbf{Training Objective.}
We adopt the \(v\)-parameterization objective \cite{salimans2022progressive}, which predicts the velocity rather than noise or clean signal. The velocity target is defined as:
\[
\mathbf{v}_t = \sqrt{\bar{\alpha}_t} \boldsymbol{\epsilon} - \sqrt{1 - \bar{\alpha}_t} \mathbf{z}_0
\]
where \(\mathbf{z}_0\) is the clean latent code, \(\boldsymbol{\epsilon} \sim \mathcal{N}(\mathbf{0}, \mathbf{I})\) is Gaussian noise, and \(\bar{\alpha}_t\) is the cumulative product of noise schedule coefficients at timestep \(t\).

The noisy latent at timestep \(t\) is computed as:
\[
\mathbf{z}_t = \sqrt{\bar{\alpha}_t} \mathbf{z}_0 + \sqrt{1 - \bar{\alpha}_t} \boldsymbol{\epsilon}
\]

Given \(\mathbf{v}_t\), we can recover both the noise and clean signal:
\[
\begin{aligned}
\mathbf{z}_0 &= \sqrt{\bar{\alpha}_t} \mathbf{z}_t - \sqrt{1 - \bar{\alpha}_t} \mathbf{v}_t \\
\boldsymbol{\epsilon} &= \sqrt{\bar{\alpha}_t} \mathbf{v}_t + \sqrt{1 - \bar{\alpha}_t} \mathbf{z}_t
\end{aligned}
\]

\textbf{Loss Function.} The training objective is:
\[
\mathcal{L}_{diff} = \mathbb{E}_{\mathbf{z}_0, t, \boldsymbol{\epsilon}, y} \left[ \lambda_t \|\mathbf{v}_t - \mathbf{v}_\theta(\mathbf{z}_t, t, y)\|_2^2 \right]
\]

where \(t \sim \mathcal{U}(1, T)\) is uniformly sampled from \([1, 1000]\), and \(\lambda_t\) is a weighting term.

\textbf{Min-SNR Loss Weighting.} We employ minimum signal-to-noise ratio (SNR) loss weighting \cite{hang2023min-snr} to balance learning across different noise levels:
\[
\lambda_t = \min\left(\frac{\bar{\alpha}_t}{1 - \bar{\alpha}_t}, \gamma\right)
\]
where \(\gamma = 5\) is a hyperparameter that prevents over-weighting low-noise timesteps.

\textbf{Noise Schedule.} We use a linear beta schedule:
\[
\beta_t = \beta_{\min} + \frac{t}{T}(\beta_{\max} - \beta_{\min})
\]
with \(T = 1000\) total diffusion steps, \(\beta_{\min} = 0.0001\), and \(\beta_{\max} = 0.02\).

\textbf{Sampling Procedure.}
For generation, we employ Denoising Diffusion Implicit Models (DDIM) \cite{song2020denoising}, which enables high-quality sampling with fewer steps than the training schedule.

Starting from pure noise \(\mathbf{z}_T \sim \mathcal{N}(\mathbf{0}, \mathbf{I})\), we iteratively denoise:
\[
\mathbf{z}_{t-1} = \sqrt{\bar{\alpha}_{t-1}} \hat{\mathbf{z}}_0 + \sqrt{1 - \bar{\alpha}_{t-1} - \sigma_t^2} \cdot \hat{\boldsymbol{\epsilon}} + \sigma_t \boldsymbol{\epsilon}_t
\]

where:
- \(\hat{\mathbf{z}}_0 = \sqrt{\bar{\alpha}_t} \mathbf{z}_t - \sqrt{1 - \bar{\alpha}_t} \hat{\mathbf{v}}_t\) is the predicted clean latent
- \(\hat{\boldsymbol{\epsilon}} = \sqrt{\bar{\alpha}_t} \hat{\mathbf{v}}_t + \sqrt{1 - \bar{\alpha}_t} \mathbf{z}_t\) is the predicted noise
- \(\sigma_t = \eta \sqrt{(1 - \bar{\alpha}_{t-1})/(1 - \bar{\alpha}_t)} \sqrt{1 - \bar{\alpha}_t/\bar{\alpha}_{t-1}}\) controls stochasticity
- \(\eta = 1.0\) determines the interpolation between DDIM (\(\eta = 0\)) and DDPM (\(\eta = 1\))
- \(\boldsymbol{\epsilon}_t \sim \mathcal{N}(\mathbf{0}, \mathbf{I})\) is fresh noise

We use \(S = 900\) sampling steps (out of \(T = 1000\) training steps) for generation, with the timestep schedule:
\[
\{t_s\}_{s=0}^{S} = \left\{\left\lfloor \frac{sT}{S} \right\rfloor\right\}_{s=0}^{S}
\]

\textbf{Training.}
The diffusion model contains approximately 10M parameters and is trained for 100 epochs using the AdamW optimizer with learning rate \(1 \times 10^{-4}\), \(\beta_1 = 0.9\), \(\beta_2 = 0.999\), and weight decay of 0.01. We use gradient clipping at norm 0.5 and a batch size of 64. The FunPhase autoencoder is frozen during diffusion training. All experiments are conducted on 4 NVIDIA A100 GPUs using PyTorch Lightning with mixed precision (FP16) training.

\subsection{\label{app:datasets} Datasets}
Our model is skeleton-agnostic and can be trained on datasets featuring a variety of characters. To demonstrate robustness and generalization, we evaluate it on both human and animal motion datasets.
All sequences are divided into 60-frame windows (1 second each), unless otherwise specified.
\begin{itemize}
\item \textsc{100STYLE} \cite{mason2022_100style}: Contains 100 distinct performative locomotion styles and over 4 million motion capture frames.
\item \textsc{Dog} \cite{zhang2018mode}: Dog motion capture dataset covering various locomotion modes such as walk, pace, and trot, as well as sitting, standing, and jumping, totaling approximately 30 minutes of motion. Given the high variation in modes, the transitions between modes represent a good challenge in terms of diverse non-periodic actions.
\item \textsc{Truebones ZOO} \cite{truebones}: A diverse collection of 70 animal skeletons, including quadrupeds, bipeds, insects, and flying birds, totaling 1,219 sequences (147,178 frames). We use the same preprocessing as \cite{gat2025anytop} and we divided sequences into 30-frame windows (1 second each).
\end{itemize}

\subsection{\label{app:metrics} Metrics}
Evaluating motion generation and reconstruction is notoriously challenging due to the need to jointly assess spatial accuracy, temporal coherence, physical plausibility, and perceptual realism \cite{zhu2023human}. Following standard practice, we employ a diverse set of metrics to ensure a fair and comprehensive evaluation. Visual examples are provided in the supplemental video.
As \textbf{reconstruction metrics}, we use:
\begin{itemize}
    \item the \emph{Joint Position Error (L2)}, measuring the average Euclidean distance between predicted and ground-truth joint positions;
    \item the \emph{Geodesic Rotation Error} (Equation 4 in main paper), the mean angular distance between predicted and target joint rotations;
    \item the \emph{Normalized Power Spectrum Similarity (NPSS)} \cite{gopalakrishnan2019neural}, comparing the frequency-domain spectra of predicted and reference motions.
\end{itemize}

We evaluate the \textbf{physical plausibility} of generated motion using:
\begin{itemize}
    \item \emph{Foot Sliding:} average velocity of grounded feet, measuring temporal drift artifacts;
    \item \emph{Foot Penetration:} the mean vertical penetration of the feet below the ground plane;
    \item \emph{Average Curve Length (ACL):} mean joint acceleration magnitude, assessing temporal smoothness;
    \item \emph{Coherence:} we measure motion coherence using a normalized smoothness score that relates internal joint motion to global locomotion. This metric penalizes excessive joint motion relative to global translation and thus reflects the smoothness and coordination of the generated motion.
\end{itemize}

To assess \textbf{generation quality}, we use a pretrained motion classifier from \cite{chen2024taming} to compute the \emph{Fréchet Inception Distance (FID)} between real and generated feature distributions, and classification \emph{Accuracy} to measure the alignment of generated motions with the intended class condition. Finally, the \emph{Diversity} is quantified as the average standard deviation among generated pose vectors across multiple samples per condition.

On the \textsc{Truebones ZOO} dataset \cite{truebones}, we follow the evaluation protocol of AnyTop \cite{gat2025anytop}: 
\begin{itemize}
    \item \emph{Coverage} quantifies how much of the real motion distribution is captured by the generated samples;
    \item \emph{Local Diversity} measures the average distance between short windows in generated motions and their nearest neighbors in the ground truth;
    \item \emph{Inter Diversity} captures variation across different generated motions, while \emph{Intra Diversity Diff} reports the difference in within-motion diversity between generated and ground-truth sequences.
\end{itemize}

\section{\label{app:ablation} Ablations}
\label{sec:supp_ablations}
In this section, we show the ablation studies that guided our final model design.

\begin{table}[h]
\centering
\caption{\textbf{Ablation comparing functional and periodic components.}}
\begin{tabular}{llccccc}
\toprule
Setting & Model & Joint Pos $\downarrow$ & Rot $\downarrow$ & Root $\downarrow$ & Sliding $\downarrow$ & Penetration $\downarrow$ \\
\midrule\midrule
No Func, Periodic & DeepPhase & 92.06 & 0.15 & 0.44 & 0.69 & 1.07 \\
Func, No Periodic & Func AE & 0.74 & 0.041 & 0.41 & 0.052 & 0.028 \\
Func, Periodic & FunPhase & \textbf{0.36} & \textbf{0.31} & \textbf{0.17} & \textbf{0.027} & \textbf{0.016} \\
\bottomrule
\end{tabular}

\label{tab:ablation_FuncVSPeriodic}
\end{table}
\subsection{Effect of functional vs periodic components}
First of all, in Table~\ref{tab:ablation_FuncVSPeriodic} we disentangle the contribution of the functional and periodic representation using the results from the experiment in Table~\ref{tab:recon}. 
Moving from DeepPhase to a purely functional autoencoder (Function AE) yields a substantial improvement in reconstruction accuracy and temporal stability, demonstrating the benefit of the functional formulation. 
Adding the periodic decomposition on top of the functional model (FunPhase) further reduces foot sliding and penetration, indicating that the two components play complementary roles: the functional formulation drives reconstruction fidelity, while phase decomposition enforces physical consistency.

\subsection{FunPhase}
\label{sec:ablation_funphase}
We trained the ablation models using $20\%$ of the \textsc{100Style} training set for $60$ epochs.

\begin{table}[h]
    \centering
    \caption{\textbf{FunPhase Ablation.} We test the performance of different architeture choices. The model 256 Channels represents our final model.}
    \begin{tabular}{lcccc}
    \toprule
         & Joint Positions & Rotation & Root Position & Number of parameters\\ \midrule\midrule
    Unified      &  3.37 &  0.42 & 0.92 & 26.1 M\\
    No FK Loss   &  26.8 &  0.097& \textbf{0.65} & 34.2 M \\
    32 Channels  &  5.68 &  0.65& 2.11 & 6.6 M\\
    128 Channels &  4.01 &  0.52& 1.57 & 17.5 M \\
    \textbf{256 Channels} &  \textbf{3.15}&  \textbf{0.092}& 0.86 & 34.2 M \\
    \bottomrule
    \end{tabular}
    
    \label{tab:PFAE_Ablations}
\end{table}

\textbf{Model size.}
We first ablate the number of latent channels in FunPhase. \autoref{tab:PFAE_Ablations} reports the reconstruction errors for models with $32$, $128$, and $256$ channels. As expected, the largest variant achieves the best performance across all metrics. Notably, even this configuration remains extremely compact, using only $256 \times 4$ values per motion clip.

\textbf{Root position encoding.}
We also evaluate a unified architecture that uses a single encoder to handle both joint rotations and the root position. In this setup, the root is treated as an additional node with a zero-valued graph positional encoding. We train this model with $256$ latent channels and refer to it as \emph{Unified}. As shown in \autoref{tab:PFAE_Ablations}, our final model (with separate encoders and decoders) consistently outperforms \emph{Unified} across all metrics. Moreover, keeping the root and joint inputs separate allows finer control over computational resources allocated to global versus local information.

\textbf{Forward Kinematic Loss.}
Finally, we assess the impact of the forward kinematic loss in training. Removing this term (\emph{No FK Loss}) results in a substantial drop in performance, as reported in \autoref{tab:PFAE_Ablations}. This confirms that the forward kinematic loss is essential for reconstructing physically plausible motion.

\subsection{Effect of Loss Components}
\label{sec:ablation_losses}

We further investigate the contribution of each term in the composite loss function (Equation~\ref{eq:total-loss} of the main paper). Specifically, we evaluate the impact of the forward kinematic loss ($\mathcal{L}_{FK}$) and the foot penalties ($\mathcal{L}_{foot}$, comprising foot sliding and foot penetration). All models are trained under the same protocol as in Appendix~\ref{sec:ablation_funphase}, using $20\%$ of the 100STYLE training set for 60 epochs.

\begin{table}[h]
\centering
\caption{\textbf{Effect of loss components.} Ablation of the forward kinematic loss ($\mathcal{L}_{FK}$) and the foot penalties (foot sliding and foot penetration). Removing $\mathcal{L}_{FK}$ leads to a sharp degradation in joint position accuracy, while removing the foot penalties results in a smaller but consistent loss in physical plausibility.}
\begin{tabular}{llccccc}
\toprule
Setting & Model & Joint Pos $\downarrow$ & Rot $\downarrow$ & Root $\downarrow$ & Sliding $\downarrow$ & Penetration $\downarrow$ \\
\midrule\midrule
Func.\ + Periodic + FK loss & FunPhase & 3.14 & \textbf{0.0092} & 0.86 & \textbf{0.078} & \textbf{0.00005} \\
Func.\ + Periodic, No FK loss & FunPhase & 26.8 & 0.097 & \textbf{0.65} & 0.40 & 0.00027 \\
Func.\ + Periodic, No FS \& FP & FunPhase & \textbf{2.88} & 0.088 & 0.89 & \textbf{0.078} & \textbf{0.00005} \\
\bottomrule
\end{tabular}

\label{tab:ablation_losses}
\end{table}

These results highlight the complementary role of the two auxiliary losses. The forward kinematic loss is critical for accurate global position reconstruction: removing it causes a roughly $9\times$ increase in joint position error and substantially worse foot sliding and penetration, confirming that explicit supervision in Cartesian space is essential to anchor the predicted rotations to physically consistent joint locations. In contrast, the foot penalties provide a smaller but consistent improvement in physical plausibility, particularly in reducing sliding artifacts, while having limited impact on overall reconstruction accuracy. Together, these components form a balanced training objective that jointly enforces geometric fidelity and physical consistency.

\begin{table*}[h]
\centering
\caption{\textbf{Input representation comparison on the \textsc{100Style} dataset.}
Representing motions using joint rotations and the root position yields consistently better performance than using global joint positions across almost all evaluation metrics.}
\small
\begin{tabular}{clcccccc}
\toprule
Representation & Method & Position (cm) $\downarrow$ & Orientation $\downarrow$ & NPSS 
$\downarrow$ & Sliding $\downarrow$ & Penetration $\downarrow$ & ACL $\downarrow$ \\
\midrule\midrule
\multirow{2}{*}{\makecell{\textsc{Global} \\ \textsc{Positions}}} & Function AE{\footnotesize-128C} & 1.82 & 0.24 & 1.95 & 0.24 & 0.004 & 1.371 \\
\phantom{Dataset} & {FunPhase}{\footnotesize-32C} & 3.66 & 0.26 & 2.17 & 0.48 & 0.037 & \textbf{1.346} \\
\phantom{Dataset} & {FunPhase}{\footnotesize-128C} & 2.99 & 0.25 & 2.05 & 0.37 & 0.022 & \underline{1.369}\\
\midrule
\multirow{2}{*}{\makecell{\textsc{Rotations} + \\ \textsc{Root Position}}} & Function AE{\footnotesize-256C} & \underline{0.74} & \textbf{0.05} & \textbf{0.59} & 0.23 & \textbf{0.001} & 1.383 \\
\phantom{Dataset} & {FunPhase}{\footnotesize-32C} & 1.93 & \underline{0.10} & 1.18 & \textbf{0.16} & \textbf{0.001} & {1.371} \\
\phantom{Dataset} & {FunPhase}{\footnotesize-256C} & \textbf{0.36} & 0.20 & \underline{0.75} & \underline{0.21} & \textbf{0.001} & {1.378}  \\
\bottomrule
\end{tabular}
\label{tab:GlobalPosAbl}
\end{table*}

\subsection{Motion representation}
Previous work has shown that directly predicting global joint positions can be advantageous for generative models \cite{meng2025absolute}. However, for our model we adopt joint rotations together with the root position as the motion representation. This choice avoids skeletal deformations and is directly compatible with standard animation software. Given a fixed skeleton, global joint positions can be efficiently obtained through forward kinematics. In contrast, predicting global positions directly may introduce errors caused by inconsistent bone lengths across frames. Moreover, converting predicted global positions back into local joint rotations—required by common motion-capture formats such as BVH—necessitates an inverse-kinematics optimization step, which is substantially more expensive than forward kinematics.

This choice is further supported by the results of our ablation in \autoref{tab:GlobalPosAbl}. We trained both FunPhase and Function AE on the full \textsc{100Style} dataset using the global joint positions, reducing the latent channels to $128$ to match the smaller dimensionality of this representation. We also present results using $32$ latent channels in both cases, to ensure a fair comparison. The results show that jointly predicting rotations and root position achieves lower reconstruction error, indicating that global joint coordinates are not advantageous for our task.

\begin{figure}
    \centering
    \begin{subfigure}{.45\textwidth}
    \centering
        \vspace{3mm}
        \begin{overpic}[scale=0.197,unit=1mm, trim=48mm 5mm 40mm 15mm,clip, grid=false]{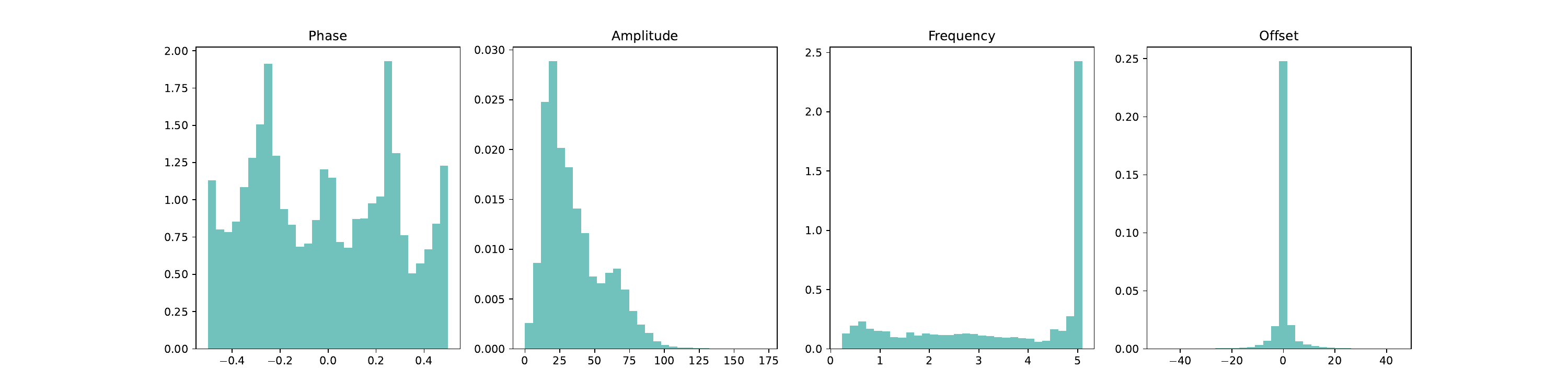}
    \put(13,26.5){\footnotesize $s_c$}
    \put(38,26.5){\footnotesize $a_c$}
    \put(62,26.5){\footnotesize $f_c$}
    \put(86,26.5){\footnotesize $b_c$}
    \end{overpic}
        \caption{Before}
    \end{subfigure}
    \hspace{0.6cm}
    \begin{subfigure}{.45\textwidth}
        \vspace{3mm}
        \begin{overpic}[scale=0.197,unit=1mm, trim=48mm 5mm 40mm 15mm,clip, grid=false]{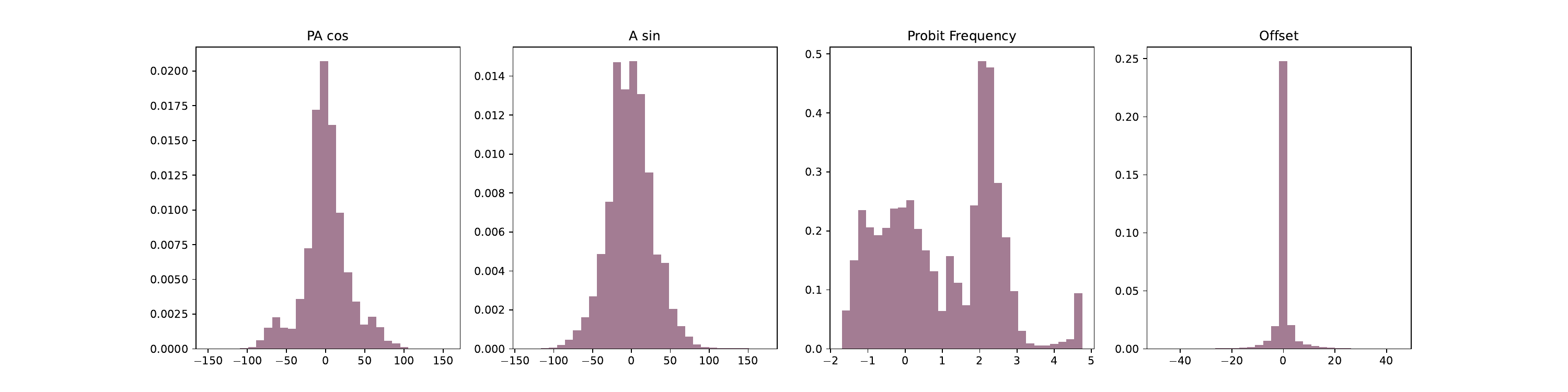}
    \put(13,26.5){\footnotesize $\mathbf{a}_{c}^{\cos}$}
    \put(38,26.5){\footnotesize $\mathbf{a}_{c}^{\sin}$}
    \put(62,26.5){\footnotesize $f_c^{probit}$}
    \put(86,26.5){\footnotesize $b_c$}
        \end{overpic}
    \caption{After}
    \end{subfigure}
    \caption{\textbf{Phase Transformation.} We plot the distribution of the latent periodic parameterization before and after the phase transformation applied in the latent diffusion model.}
    \label{fig:phaseTransform}
\end{figure}

\begin{table}[ht]
    \centering
    \caption{\textbf{Phase Transformation Ablation.}
Comparison between the Diffusion Latent Model with and without Phase Transformation.}
    \begin{tabular}{lccc}
    \toprule
    \makecell{Phase \\ Transf.} & FID $\downarrow$ & Accuracy (\%) $\uparrow$ & Diversity $\uparrow$ \\ 
    \midrule\midrule
    No  & $1.28\textsuperscript{\scriptsize $\pm$0.44}$ & 34.83 & $\mathbf{0.56}\textsuperscript{\scriptsize $\pm$0.01}$\\
    Yes  & $\mathbf{1.27}\textsuperscript{\scriptsize $\pm$0.48}$ & \textbf{37.6} & $\mathbf{0.56}\textsuperscript{\scriptsize $\pm$0.01}$\\
    \bottomrule
    \end{tabular}
    \label{tab:PT_results}
\end{table}

\subsection{\label{app:phaseTransf}Phase Transformation}
The periodic parameterization, while compact and interpretable, poses challenges for diffusion modeling due to the domain and distribution of its parameters. To address this, we apply domain transformations that make the representation more compatible with Gaussian diffusion. \autoref{fig:phaseTransform} illustrates how the Phase Transformation described in Equation \ref{eq:phase-transform} of the main manuscript reshapes the distribution of the latent phase parameters. The resulting distributions more closely resemble a Normal distribution, which facilitate the diffusion process .

To evaluate its benefits, we train Latent Diffusion models (with and without the Phase Transformation) using 2M parameters on a $20\%$ subset of the \textsc{100Style} training set. As shown in \autoref{tab:PT_results}, adding the Phase Transformation improves performance under these conditions. Furthermore, the effectiveness of phase parameterization in latent diffusion is reinforced by the substantial performance gains observed when comparing Function Diffusion with our FunPhase Diffusion in \autoref{tab:diffusion} of the main paper.

\section{\label{app:phaseManifold} Learned Phase Manifold.}
\begin{figure}[ht]
    \centering
    \hfill
    \begin{subfigure}{.3\textwidth}
        \centering
        \includegraphics[width=.95\linewidth]{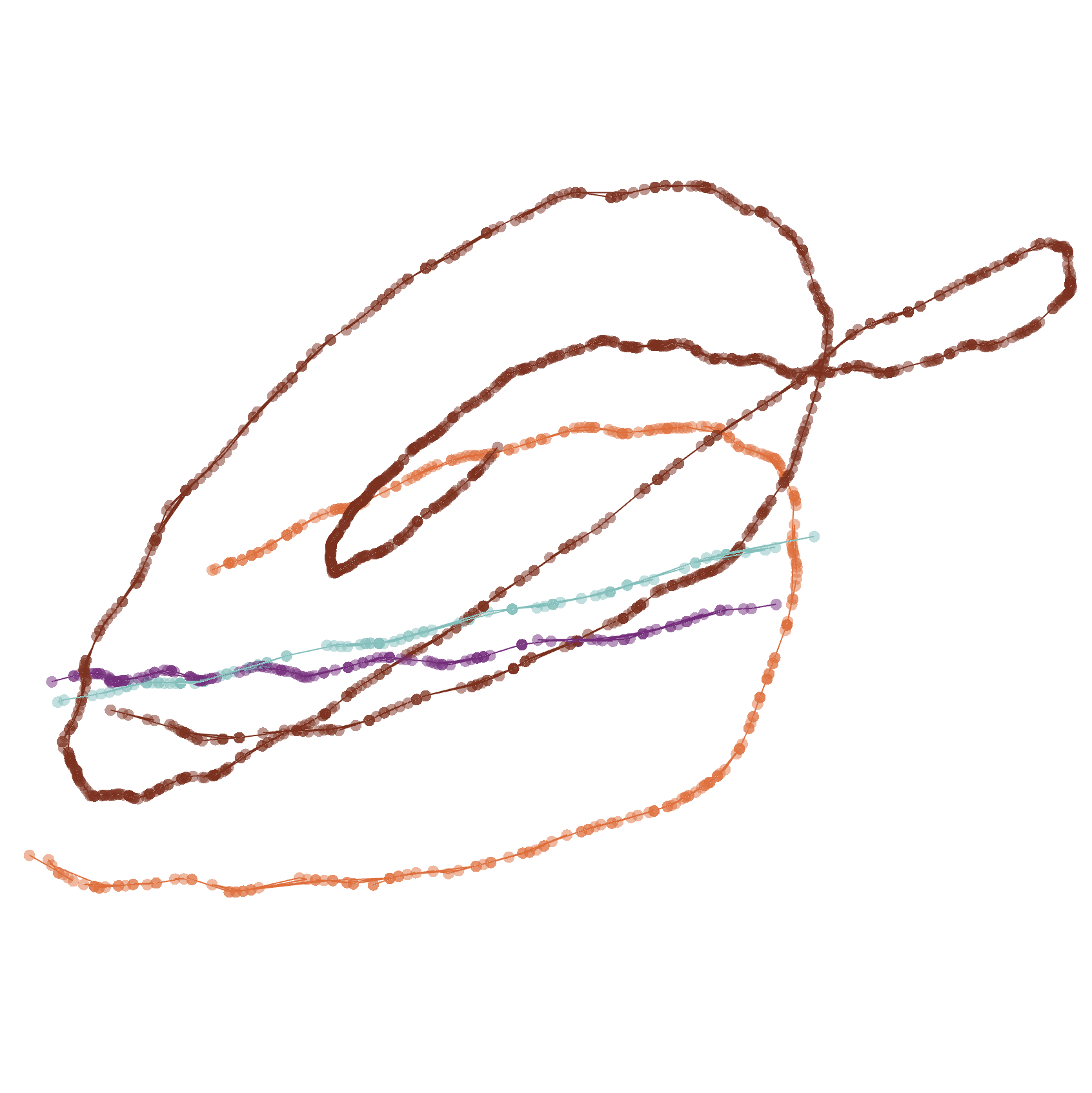}  
        \caption{Input space}
    \end{subfigure}
    \hfill
    \begin{subfigure}{.3\textwidth}
        \centering
        \includegraphics[width=.95\linewidth]{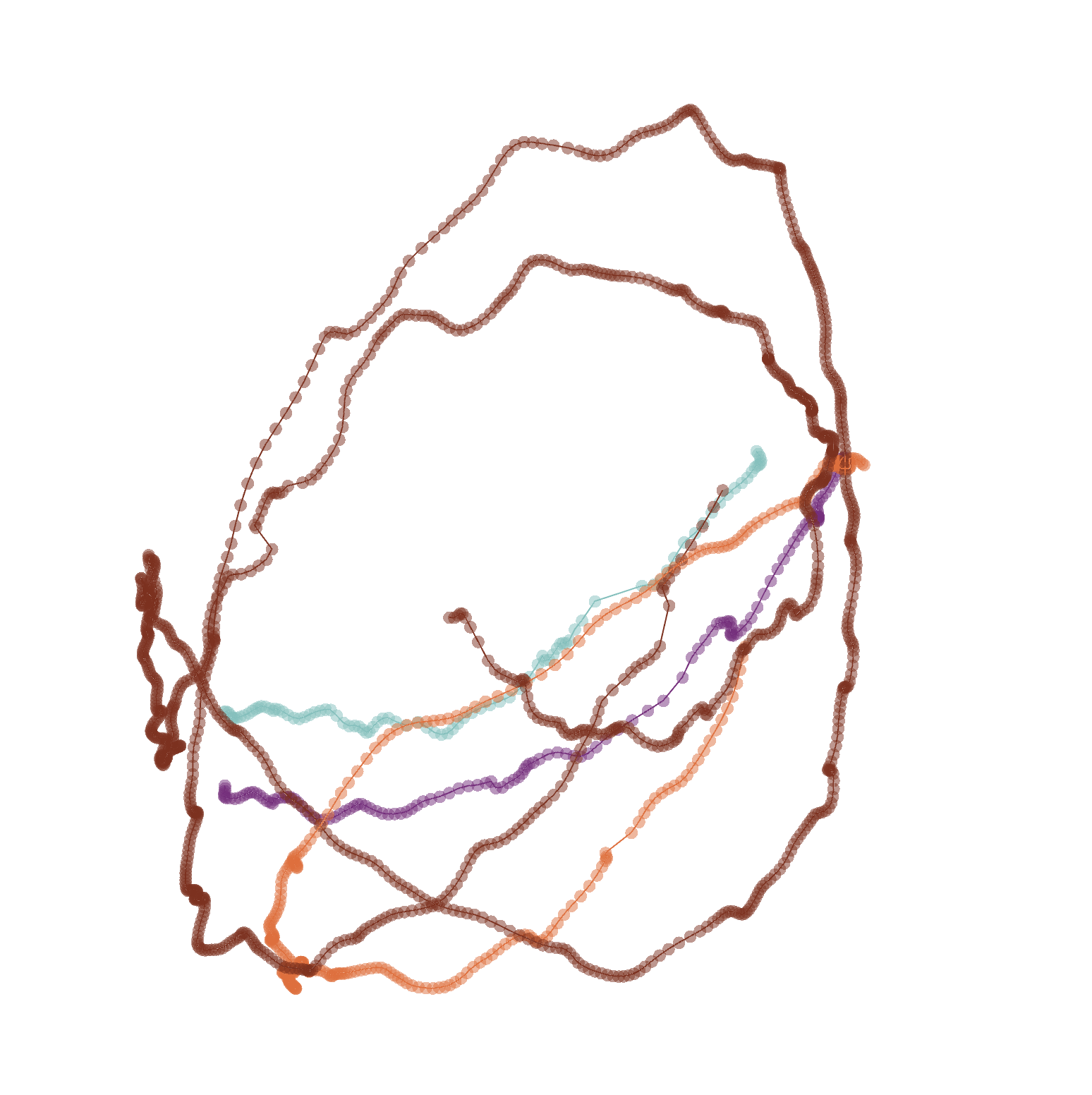}  
        \caption{DeepPhase}
    \end{subfigure}
    \hfill
    \begin{subfigure}{.3\textwidth}
        \centering
        \includegraphics[width=.95\linewidth]{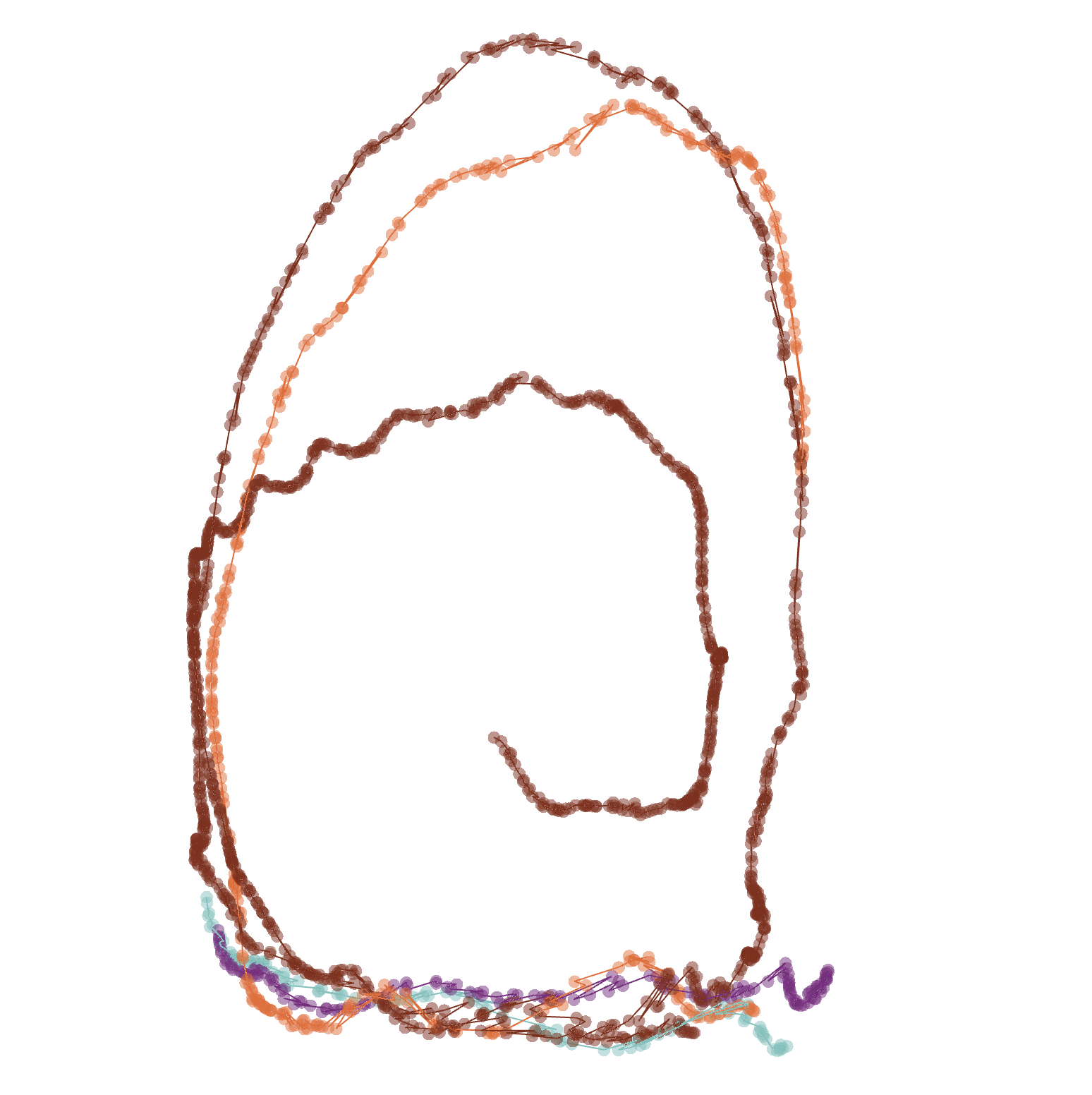}  
        \caption{FunPhase}
    \end{subfigure}
    \hfill
    \caption{\textbf{Phase Manifold.} The plots show the phase manifolds obtained with DeepPhase and FunPhase, alongside the original motion features. All encoded sequences correspond to a dog-running motion.}
    \label{fig:phases}
\end{figure}

We visualize the learned phase manifold and compare it against the one obtained with DeepPhase.
After computing the phase manifold on running sequences from the \textsc{Dog} dataset, we extract the principal components (PCs) of the phase features and project them onto a 3D plane (\autoref{fig:phases}).
For comparison, we compute the phase embeddings from DeepPhase and similarly project their PCs to 3D.
We also visualize the PCs of the original motion features (root position and joint angles).
We observe that our phase representation maintains a compact circular structure characteristic of cyclic motion, whereas the original motion features collapse into disorganized linear trajectories.
This structured phase representation is known to be effective for downstream applications such as motion matching and control  \cite{starke2022deepphase}.

\section{Performance Analysis under Motion Aperiodicity}
\label{app:aperiodicity}

To better characterize the limitations introduced by phase-based representations, we analyze model performance as a function of motion
aperiodicity. We quantify aperiodicity via the spectral entropy of each motion
sequence and partition the test set into quartiles of increasing entropy.
Reconstruction and physical-plausibility metrics are reported per quartile in
Table~\ref{tab:entropy_analysis}.

\begin{table}[h]
\centering
\caption{\textbf{Reconstruction and physical-plausibility metrics across
spectral entropy quartiles.} Higher entropy corresponds to more aperiodic
motion. Means and standard deviations are reported across all sequences in
each bin.}
\label{tab:entropy_analysis}
\small
\setlength{\tabcolsep}{4pt}
\begin{tabular}{lcccccc}
\toprule
Entropy Group & Entropy Range & Position $\downarrow$ & Orientation $\downarrow$ & Foot Sliding $\downarrow$ & Foot Penetration $\downarrow$ & ACL $\downarrow$ \\
\midrule\midrule
Lowest    & $[0.176,\,0.384]$ & $0.067 \pm 0.23$ & $0.12 \pm 0.088$ & $0.49 \pm 0.42$  & $6.29\mathrm{e}{-7} \pm 7.48\mathrm{e}{-7}$ & $1.18 \pm 1.02$ \\
Low Mid   & $[0.384,\,0.437]$ & $0.087 \pm 0.25$ & $0.14 \pm 0.10$  & $0.55 \pm 0.53$  & $6.62\mathrm{e}{-7} \pm 6.54\mathrm{e}{-7}$ & $1.32 \pm 1.13$ \\
High Mid  & $[0.437,\,0.504]$ & $0.11 \pm 0.51$  & $0.15 \pm 0.145$ & $0.65 \pm 0.81$  & $7.73\mathrm{e}{-7} \pm 6.44\mathrm{e}{-7}$ & $1.62 \pm 1.14$ \\
Highest   & $[0.504,\,0.973]$ & $0.062 \pm 0.25$ & $0.11 \pm 0.097$ & $0.662 \pm 0.439$ & $1.09\mathrm{e}{-6} \pm 6.59\mathrm{e}{-7}$ & $2.19 \pm 0.99$ \\
\bottomrule
\end{tabular}
\end{table}

\paragraph{Quantitative trends.} Increasing aperiodicity primarily affects
\emph{temporal coherence} and \emph{contact consistency}: ACL nearly doubles
($1.18 \rightarrow 2.19$) and foot sliding worsens ($0.49 \rightarrow 0.66$)
as we move from the lowest- to the highest-entropy bin, indicating less
smooth and less stable motion. In contrast, the reconstruction errors
(position and orientation) do not degrade monotonically with entropy,
suggesting that FunPhase can still reconstruct aperiodic motions reasonably
well, even though its inductive bias is rooted in periodic structure.

\paragraph{Qualitative content per bin.} A caption-level analysis of the
sequences in each quartile reveals a consistent progression in motion content:
\begin{itemize}
    \item \textbf{Low entropy:} locomotion-dominated motions
          (e.g., walking, simple trajectories);
    \item \textbf{Low-mid entropy:} predominantly locomotion but with greater
          variability (e.g., direction changes, jogging);
    \item \textbf{High-mid entropy:} more dynamic full-body actions
          (e.g., jumping, dancing);
    \item \textbf{Highest entropy:} interaction- and upper-body-driven motions
          (e.g., clapping, gesturing, transitions between heterogeneous actions).
\end{itemize}

Overall, spectral entropy correlates with motion complexity and temporal
structure rather than with specific action categories. The trends in
Table~\ref{tab:entropy_analysis} are consistent with the discussed limitations: phase-based modeling is most accurate on cyclic, locomotion-like
motions and gradually loses temporal-smoothness fidelity on weakly periodic
or transition-heavy sequences, even when frame-wise reconstruction remains
strong.

\begin{table}[h]
\caption{\textbf{Reconstruction error under decreasing keyframe distances.}
    Given a fixed window of 50 frames, each model reconstructs the full motion from a subsampled set of keyframes. FunPhase provides more accurate interpolation than both standard autoencoders and linear interpolation (SLERP) at shorter intervals.}
    \centering
    \begin{tabular}{lcccc}
\toprule
    KF Dist. & 25 & 15 & 10 & 5 \\ \midrule\midrule
    SLERP & \textbf{14.79} & \textbf{4.11} & 1.32 & 0.43 \\
    Function AE & 126.21 & 6.02 & 1.29 & 0.76\\ 
    FunPhase & 151.24 &  11.61 & \textbf{0.99} & \textbf{0.39} \\  
\bottomrule
    \end{tabular}
    
    \label{tab:keyframesAE}
\end{table}

\section{Reconstruction error under increasing keyframe distances.}
\label{app:increasing_dist}

In \autoref{tab:keyframesAE}, we evaluate the reconstruction performance of our model when conditioned on keyframes sampled at increasing temporal intervals. We compare Function AE and FunPhase against a SLERP baseline, which linearly interpolates the root trajectory and joint rotations between keyframes. FunPhase shows a substantial improvement at a keyframe distance of $10$ and nearly matches full-sequence reconstruction performance at a distance of $5$. 
At higher distances, the performance dropped. This could be due to the fact the model never saw big gaps between input keyframes. We believe that performing a training with more sparse input could further improve the performance of our model.

\end{document}